\documentclass[runningheads]{llncs}

\usepackage{eccv}

\usepackage{eccvabbrv}

\usepackage{graphicx}
\usepackage{booktabs}

\usepackage{bm}
\usepackage{times}
\usepackage{epsfig}
\usepackage{float} 
\usepackage{makecell}
\usepackage{colortbl}
\usepackage{mathtools}
\usepackage{extarrows}
\usepackage{algorithmic}
\usepackage{adjustbox}
\usepackage[percent]{overpic}
\usepackage[linesnumbered,ruled]{algorithm2e}
\SetKwComment{Comment}{$\triangleright$\ }{}
\usepackage{balance}
\usepackage{stfloats}
\usepackage{textcomp}
\usepackage{subcaption}
\usepackage{pifont}
\usepackage{enumitem}

\usepackage{tcolorbox}
\tcbuselibrary{listings, breakable, skins}
\usepackage{listings}
\usepackage{wrapfig}

\usepackage{pifont}
\newcommand{\cmark}{\ding{51}}
\newcommand{\xmark}{\ding{55}}

\usepackage{xcolor}

\definecolor{darkgray}{RGB}{80,80,80}
\definecolor{lightgray}{RGB}{245,245,245}

\usepackage{multirow}
\usepackage{comment}

\definecolor{MyBlue}{RGB}{33,85,205}
\definecolor{gray}{HTML}{E5E4E2} 
\definecolor{cGreen}{HTML}{B7D5A6} 
\definecolor{cBlue}{HTML}{EFFFFF} 
\definecolor{cOrange}{HTML}{F8C794}

\graphicspath{{figures/}}

\usepackage[accsupp]{axessibility}  

\usepackage{hyperref}

\usepackage{orcidlink}

\begin{document}

\title{DocArena: Turning Raw Documents into Controllable Training Environments for Document Search Agents}

\newcommand{\tong}[1]{\textcolor{green}{#1}}
\newcommand{\samya}[2]{\textcolor{orange}{#2}}

\titlerunning{DocArena}

\author{Jiamian Wang\inst{1}\thanks{Work done during
an internship at Adobe Research.} \and
Ruyi Zhang\inst{2} \and
Tong Yu\inst{2} \and 
Jing Shi\inst{2} \and 
Samyadeep Basu\inst{2} \and 
Rajiv Jain\inst{2} \and 
Zhiqiang Tao\inst{1} \and 
Tong Sun\inst{2}
}

\authorrunning{J.~Wang et al.}

\institute{Rochester Institute of Technology \and
Adobe Research
}
\maketitle

\begin{abstract}
Recent methods train search agents via reinforcement learning from (question, answer, evidence) tuples without requiring expert trajectories. The tuples serve as the training environment, and whose properties directly shape what search strategies and generalization abilities the agent can develop. While prior works have made encouraging progress in improving training data quality, existing environments remain predominantly text-based and existing approaches can struggle to construct training environments that are controllable, scalable, and account for multimodal data. Given this, we propose DocArena, a fully automated data curation pipeline building on the practical need for multimodal document search and question-answering. It transforms raw document collections into training environments for search agents without any human annotation. The pipeline first structures and indexes documents through MLLM-based visual perception, then profiles and leverages the cross-page information distribution to construct reasoning-intensive QA pairs, as well as performs cascaded quality assurance operations via MLLM. We introduce DocArena-79K with QA pairs from 8,336 documents spanning 16 domains and 49 languages. We further design a Doc-Search agent infrastructure that decouples visual perception from the policy model, allowing text-based LLMs to serve as the reasoning backbone for multimodal document retrieval and QA. Under a unified evaluation framework where only the policy model differs, experiments on six multimodal document scenarios and seven text-based QA benchmarks show that agents trained on DocArena data achieve the best performance on both retrieval accuracy and QA quality. Further analysis on agent search behaviors confirms the effectiveness and controllability of the constructed training environment.

\end{abstract}

\section{Introduction}\label{sec:intro}
Search agents that interact with external retrieval systems to answer questions have attracted growing attention~\cite{jin2025search,song2025r1,chen2025learning}. By formulating search queries, incorporating retrieved results, and performing multi-step reasoning, search agents tackle  knowledge-intensive tasks across diverse information sources~\cite{zheng2025deepresearcher,guo2025deepseek}. Recent RL-based methods train such agents from (question, answer, evidence) triplets $(q, a, \mathcal{E})$ without requiring full expert trajectories. The community has advanced search agents from multiple axes, \emph{e.g.}, improving reward modeling~\cite{song2025r1,lu2025arpo}, enhancing search tools such as web search APIs~\cite{wei2025webagent,li2025webthinker} or document parsers, and designing better policy optimization~\cite{shao2024deepseekmath,yu2025dapo}.

Beyond tools and algorithms, a more fundamental factor is the \textit{training environment} itself. In RL-based post-training (\emph{e.g.}, GRPO), each $(q, a, \mathcal{E})$ triplet shapes what the agent can learn: the question $q$ defines the search task, the evidence pages $\mathcal{E}$ constrain the observable state space that the agent can interact with, and the answer $a$ provides the outcome supervision. The diversity, correctness, and structure of these triplets directly shape the search strategies, reasoning depth, and generalization ability that the agent can develop. Yet constructing the triplets is far from straightforward. For example, real-world document collections are inherently noisy and heterogeneous, with uneven information density across pages, varying difficulty levels, and practical barriers such as licensing restrictions and domain-specific formatting. A general-purpose sampling approach may not satisfy the requirements for evidence correctness, reasoning diversity, and retrieval necessity that training effective search agents demands.

Recent works on RL with verifiable rewards have shown that scaling and diversifying training environments directly improves reasoning ability~\cite{zeng2025rlve,stojanovski2025reasoning}.
ZeroSearch~\cite{sun2025zerosearch} uses an LLM to simulate search engine responses. IKEA~\cite{huang2025reinforced} applies model probing to classify samples by difficulty. R1-Searcher~\cite{song2025r1} grades difficulty via rollout statistics. However, several limitations remain. First, training data is predominantly text-based, lacking consideration for multimodal data, \emph{e.g.}, tables, charts, figures, and complex layouts in document collections. Second, existing approaches can still involve manual curation~\cite{jin2025search,shi2025search} that limits scalability to new domains and large volumes. Third, it remains difficult to precisely control key data properties, including the correctness of evidence annotations, the diversity of reasoning types, the search depth, and the domain coverage, limiting the ability to provide customized guidance for training search agents.

This work studies an under-explored problem: \textit{how to build a fully automated data curation pipeline to construct controllable, scalable, and generalizable training environments for search agents users.} Building on the practical application of multimodal document search and question-answering, we propose DocArena, an end-to-end pipeline that transforms raw document collections into training-ready (question, answer, evidence) tuples. The pipeline operates in four stages:
(1)~\textit{Document structuring and indexing} converts raw PDFs into page images, structured semantic text, and a dense retrieval index, which adopts MLLM-based visual perception to extract the semantic content of each page.
(2)~\textit{Cross-page information distribution profiling} analyzes across-page knowledge point distributions. By identifying uniquely or broadly distributed information, the pipeline performs the accurate alignment between evidence and the constructed QA pair, ensuring the answer exclusiveness and minimizing the annotation noise.
(3)~\textit{Evidence-grounded reasoning chain construction} leverages the distribution profile to select evidence pages and build reasoning-intensive QA pairs, by which the pipeline produces diverse, scalable reasoning chains through predefined templates with the diversity control.
(4)~\textit{Cascaded quality assurance} applies a multi-layer verification system from deterministic filters through MLLM-based checks, which removes noise introduced by earlier stages, provides customized data gating to shape the training environment, and improves overall data quality.

With the proposed pipeline, we construct DocArena-79K, a large-scale training dataset of $79,623$ QA pairs from $8,336$ source documents spanning $16$ content domains and $49$ languages. The dataset covers $37,480$ unique evidence pages with a total of $238,271$ page references, where $92.3\%$ of samples require multi-page evidence (ranging from $1$ to $20$ pages). Evidence pages span diverse modalities: $63.2\%$ contain at least one table, figure, or chart beyond plain text, with the top modality combinations being text-only ($36.8\%$), text+figure ($27.7\%$), and text+table ($19.9\%$). The source documents cover $10$ document types (technical reports, scientific papers, manuals, forms, etc.) with no single domain exceeding $11\%$, reflecting balanced domain coverage.

To validate the effectiveness and controllability of the constructed training environment, we train search agents on DocArena-79K. Besides, we design a Doc-Search agent infrastructure that decouples visual perception from the policy model, consisting of an LLM policy, a multimodal retriever, and an online OCR tool. This design enables effective multimodal document understanding while maintaining compatibility with existing text-based search agents. We compare against multiple search agent methods under this unified infrastructure, with the policy model as the sole controlled variable. Experiments on six multimodal document scenarios (MMLongBench-Doc single-page and multi-page, VisRBench FigureQA, TableQA, and TextQA, and SlideVQA) show that the proposed Doc-Search agent achieves the best performance on both retrieval and QA quality. On seven text-based QA benchmarks, our agent ranks first on four out of seven, demonstrating strong generalization from multimodal document environments to text-based scenarios. By analyzing the search behaviors of Doc-Search agent, we further confirm the effectiveness and controllability of the constructed environment. The contributions are summarized as follows:
\begin{itemize}
    \item We propose DocArena, a fully automated pipeline that constructs controllable, scalable, and generalizable training environments for search agents from raw multimodal document collections, without any human annotation.
    \item We curate DocArena-79K, a large-scale training dataset that covers diverse reasoning types, multi-page evidence grounding, and rich visual-structural content across multiple domains and languages.
    \item We design a Doc-Search agent infrastructure that decouples visual perception from the policy model, allowing text-based LLMs to serve as the reasoning backbone for multimodal document retrieval and QA under a unified framework.
    \item Experiments show the effectiveness of proposed search agent on both retrieval and QA performance overall. We hope the proposed pipeline may inspire future study toward dynamic, customized, and automated training environment construction for search agents on multimodal document data.
\end{itemize}

\section{Related Works}
\label{sec:related_works}

\textbf{Document Understanding.}
Documents consist of heterogeneous elements such as tables, charts, figures, and complex layouts, which require models to jointly capture semantics and structure~\cite{livathinos2025docling,ouyang2025omnidocbench,yang2025cc,li2020docbank,wu2025moloragbootstrappingdocumentunderstanding,tian2025mmcr}. Recent methods~\cite{xiao2025adaptive,wang2025marten,zhu2025simple,chen2025document,tanaka2025vdocrag,caffagni2025recurrence} have enhanced single-page modeling, \emph{e.g.}, DocLLM~\cite{wang-etal-2024-docllm} and DocLayLLM~\cite{liao2025doclayllm} improve the text–spatial alignment, while TextMonkey~\cite{liu2024textmonkey} avoids OCR errors with an end-to-end vision-based approach. These methods show strong performance on table or figure-centric inputs but remain confined to single-page settings. For long-document QA, SV-RAG~\cite{chen2024sv}, Doc-React~\cite{wu-etal-2025-doc},  MDocAgent~\cite{han2025mdocagent}, and M3DocRAG~\cite{cho2024m3docrag} explore MLLM-based self-retrieval, reasoning decomposition, or multi-agent collaboration for long-document QA. Reasoning-Table~\cite{lei2025reasoning} applies reinforcement learning to the structured table reasoning. Besides, DocDancer~\cite{zhang2026docdancer} focuses primarily on the SFT stage and leaves the post-training of agentic multimodal search and QA underexplored.

\textbf{Agentic Search.}
We study search agents~\cite{jin2025search,jin2025empirical,gao2025beyond,zhang2025router,wu2025masksearch,wu2025mmsearch,zheng2025deepresearcher,fan2025deepplanner,hu2026sagebenchmarkingimprovingretrieval,miroyan2025searcharenaanalyzingsearchaugmented,han2024ragqaarena} for multimodal, multi-page, reasoning-intensive document retrieval and QA. Despite the prosperity, leading methods can face challenges in this scenario. Among them, Search-R1~\cite{jin2025search} is among the first to apply outcome-based rewards for incentivizing multi-turn search and reasoning, primarily focusing on text-based scenarios. ZeroSearch~\cite{sun2025zerosearch} replaces online search with a generator-based simulator to control context noise and improve training stability. IKEA~\cite{huang2025reinforced} designs a knowledge-boundary aware reward to balance internal and external knowledge utilization during search. AutoRefine~\cite{shi2025search} introduces explicit knowledge refinement steps between retrieval rounds to distill and organize retrieved evidence, primarily evaluated on text-based QA benchmarks. $O^2$-Searcher~\cite{mei20252} addresses open-domain open-ended QA by combining flexible retrieval and generation strategies. Besides, R-Search~\cite{zhao2025r} leverages multi-reward reinforcement learning, designed primarily for open-domain text-based scenarios. Different from above methods, this work  fills the gap by targeting at the scenario with newly-curated training dataset and the multimodal agentic search infrastructure, achieving more effective retrieval and reasoning performance.

\textbf{Training Environment for Search Agents.}
The concept of training environments for LLM RL has been 
established in the reasoning domain, where environments are 
defined by problem generators and verifiable 
rewards~\cite{yu2025learning,stojanovski2025reasoning,yu2025learning}.
For search agents, the training environment takes the form of 
$(q, a, \mathcal{E})$ triplets that define search tasks and 
reward signals.
Recent search agent post-training methods have benefited from improving the training environment, \emph{e.g.}, IKEA~\cite{huang2025reinforced} classifies samples into easy/hard splits via model probing, teaching agents when to search. R1-Searcher~\cite{song2025r1} grades difficulty by rollout count and stages training accordingly. ZeroSearch~\cite{sun2025zerosearch} collects query-document pairs from real search engines to train a simulation model. REDSearcher~\cite{chu2026redsearcher} synthesizes multi-hop tasks from knowledge graph topologies with evidence dispersion control. ASearcher~\cite{gao2025beyond} iteratively augments seed questions through injection and fuzzing mechanisms. While these methods have made valuable progress, existing environment construction approaches can still involve manual processing and have not yet converged to a standardized pipeline, making it difficult to precisely construct, control, and scale the training environment to fully unlock the potential of search agents. This work introduces {DocArena}, an automated pipeline that standardizes the construction of search agent training environments directly from raw document collections, with curated dataset presented and search agent trained upon it.

\section{Method}\label{sec:method}

We present the search agent and the scope of our study in Section~\ref{subsec: preliminary}. We discuss what makes a good training environment (Section~\ref{subsec: env}), present the automated data-curation pipeline (Section~\ref{subsec: pipeline}), and describe the Doc-Search agent infrastructure (Section~\ref{subsec: agent}).

\subsection{Preliminaries}\label{subsec: preliminary}
\textbf{Search Agent.}
A search agent interacts with an external search engine to answer a given question. At each turn, the agent formulates a search query, incorporates the retrieved results, and performs reasoning. Unlike imitation learning approaches that require full expert trajectories, recent RL-based methods~\cite{shao2024deepseekmath,jin2025search,zheng2025deepresearcher} train agents from $(q, a, \mathcal{E})$ triplets without step-level supervision. This scopes our study to the training data as the environment from which the agent learns search behaviors through reward signals.

\textbf{Notations.}
We instantiate the search agent in the multimodal document setting, a more complex and practically relevant scenario where the retrieval corpus consists of visually rich PDF documents. A document $d$ consists of pages $\{p_1, \ldots, p_N\}$, and we adopt $\mathcal{P}$ for the full page collection across all documents. Each page $p$ has a rendered image $p^{\text{img}}$ and extracted metadata $p^{\text{meta}}$. A training sample is a triplet $(q, a, \mathcal{E})$ consisting of a question $q$, a ground-truth answer $a$, and an evidence page set $\mathcal{E} \subset \mathcal{P}$. During training, the agent generates rollout trajectories given $q$.

\begin{figure*}[t] 
\centering 
\includegraphics[width=\textwidth]{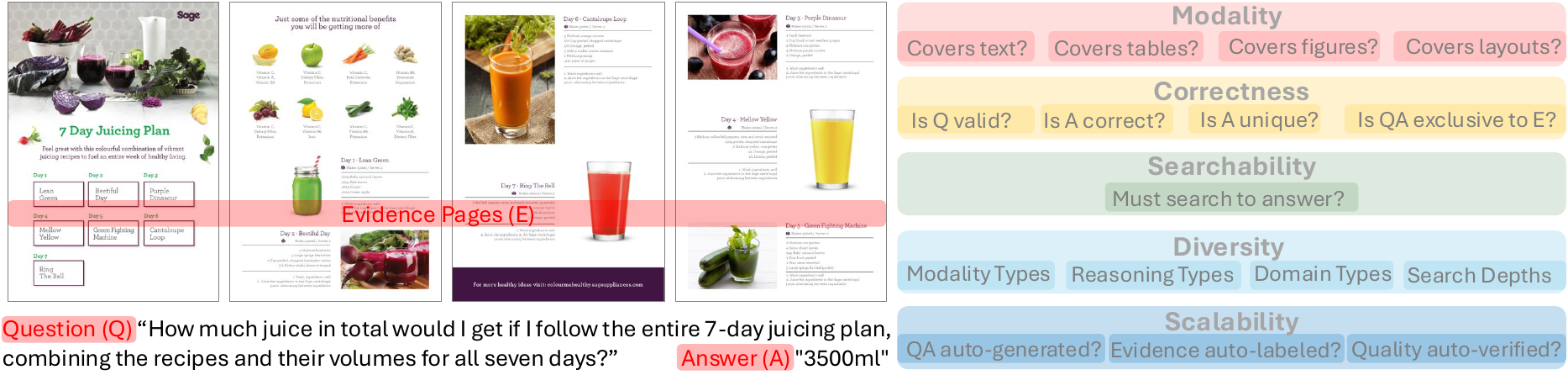} 
\vspace{-5mm}
\caption{Illustration of the training data used for RL training of search agents. Each sample is a triplet (question, answer, evidence) without intermediate trajectories (\textit{left}). Constructing a high-quality training environment is non-trivial, as it must simultaneously satisfy multiple retrieval-related quality dimensions (\textit{right}).}
\vspace{-4mm}
\label{fig: search_agent_data} 
\end{figure*}

\begin{table*}[t]
\caption{
Training data construction and data properties comparison. We include some representative search agent methods and the related QA synthesis works, assessed by the proposed five-dimensional quality model (Section~\ref{subsec: env}).
}\label{tab:env}
\vspace{-2mm}
\centering
\resizebox{\textwidth}{!}{
\begin{tabular}{l|c|l|c|cccc|c|ccc}
\hline
\multirow{2}{*}{\textbf{Method}} & \multirow{2}{*}{\textbf{Year}} & \multirow{2}{*}{\textbf{Data Source}} & \textbf{Correctness} & \multicolumn{4}{c|}{\textbf{Diversity}} & \textbf{Searchability} & \multicolumn{3}{c}{\textbf{Scalability}} \\
\cline{4-12}
& & & Exclusivity & Modality & Reasoning & Domain & Depth & Retrieval Need & QA Synth. & Evid. Label & Auto Verif. \\
\hline
\multicolumn{12}{l}{\textit{Search Agent Methods}} \\
\hline
Search-R1~\cite{jin2025search}       & 2025 & NQ + HotpotQA         & \xmark & Text & Factual     & Wiki      & 1--2 & \xmark & \xmark & \xmark & \xmark \\
AutoRefine~\cite{shi2025search}   & 2025 & NQ + HotpotQA         & \xmark & Text & Factual     & Wiki      & 1--2 & \xmark & \xmark & \xmark & \xmark \\
IKEA~\cite{huang2025reinforced}             & 2025 & NQ + HotpotQA         & \xmark & Text & Factual     & Wiki      & 1--2 & Partial & \xmark & \xmark & \xmark \\
R1-Searcher~\cite{song2025r1} & 2025 & HotpotQA + 2Wiki      & \xmark & Text & Multi-hop     & Wiki      & 2    & \xmark & \xmark & \xmark & Partial \\
ReSearch~\cite{chen2025learning}          & 2025 & MuSiQue               & \xmark & Text & Multi-hop     & Wiki      & 2--4 & \xmark & \xmark & \xmark & \xmark \\
ZeroSearch~\cite{sun2025zerosearch}   & 2025 & NQ + HotpotQA + Sim   & \xmark & Text & Multi-hop     & Wiki      & 1--2 & \xmark & Partial & \xmark & \xmark \\
SWiRL~\cite{goldie2025synthetic}          & 2025 & HotpotQA / GSM8K      & \xmark & Text & Mixed       & Wiki+Math & 1--2 & \xmark & \cmark & \xmark & Partial \\
REDSearcher~\cite{chu2026redsearcher}    & 2026 & KG-DAG Synthesis      & \cmark & Text & Multi-hop  & Wiki+Web  & 2--6 & \cmark & \cmark & \cmark & \cmark \\
ASearcher~\cite{gao2025beyond}        & 2025 & Seed QA Augment.      & \xmark & Text & Mixed       & Wiki      & 2--4 & \cmark & \cmark & \xmark & \cmark \\
\hline
\multicolumn{12}{l}{\textit{QA Data Synthesis (not for agent training)}} \\
\hline
GRADE~\cite{lee2025grade}             & 2025 & Document KG$\to$QA    & \cmark & Text & Multi-hop   & Multi     & 2--4 & -- & \cmark & \cmark & \cmark \\
MultiHop-RAG~\cite{tang2024multihop} & 2024 & News Articles       & \xmark & Text & Multi-hop   & News      & 2--4 & -- & \cmark & Partial & Partial \\
RARE~\cite{zeng2025rare}                  & 2025 & Custom Corpus         & \xmark & Text & Relational  & Multi     & 1--3 & -- & \cmark & Partial & Partial \\
\hline
\rowcolor{gray!15}
\textbf{DocArena (Ours)}                         & 2026 & Raw Documents         & \cmark & Mixed & >4 Types    & Multi     & 1--20 & \cmark & \cmark & \cmark & \cmark \\
\hline
\end{tabular}
}
\vspace{-2mm}
\label{tab:data_comparison}
\end{table*}

\textbf{The Training Environment.}
Previous works advance search agents from different aspects, \emph{e.g.}, improving reward modeling with progress rewards~\cite{lu2025arpo} or trajectory-level credit assignment~\cite{song2025r1,yu2025dapo}, and enhancing search tools such as web search APIs~\cite{li2025webthinker,wei2025webagent} or document parsers. By contrast, this work studies a complementary and potentially overlooked aspect: \textit{the training data itself as the environment}. We investigate three questions: (1) What makes a good training environment for search agents? (Section~\ref{subsec: env}) (2) How to construct, control, and scale such environments? (Section~\ref{subsec: pipeline}) (3) What is the impact of the constructed environment? (Section~\ref{subsec: agent})

\subsection{What Makes a Good Environment for Search Agents?}\label{subsec: env}

We identify five quality dimensions that a search agent training environment could satisfy in Fig.~\ref{fig: search_agent_data} and compare in Table~\ref{tab:env}.
\textbf{(1) Correctness.} Beyond standard QA correctness, search agent data requires \textit{evidence exclusivity}, i.e., the answer $a$ must be derivable \textit{only} from the annotated evidence pages $\mathcal{E}$, not from other pages in the collection. Otherwise, the data introduces noise, and the environment provides incorrect feedback during training.
\textbf{(2) Diversity.} The training data should cover diverse reasoning types (comparison, conditional inference, consistency checking, etc.), diverse domains, and diverse search depths. Single-type data leads to rigid search strategies that fail to generalize.
\textbf{(3) Searchability.} Whether the base model truly needs external retrieval to answer the question, rather than relying on parametric knowledge.
\textbf{(4) Modality.} Real-world documents are visual objects containing tables, figures, charts, and layout structures. Training data should preserve these visual-structural semantics and align with the retriever modality at the page level.
\textbf{(5) Scalability.} Whether the training environment can be readily scaled to new domains and large volumes without human annotation.

\subsection{DocArena: Automated Curation Pipeline}\label{subsec: pipeline}

\begin{figure*}[t] 
\centering 
\includegraphics[width=\textwidth]{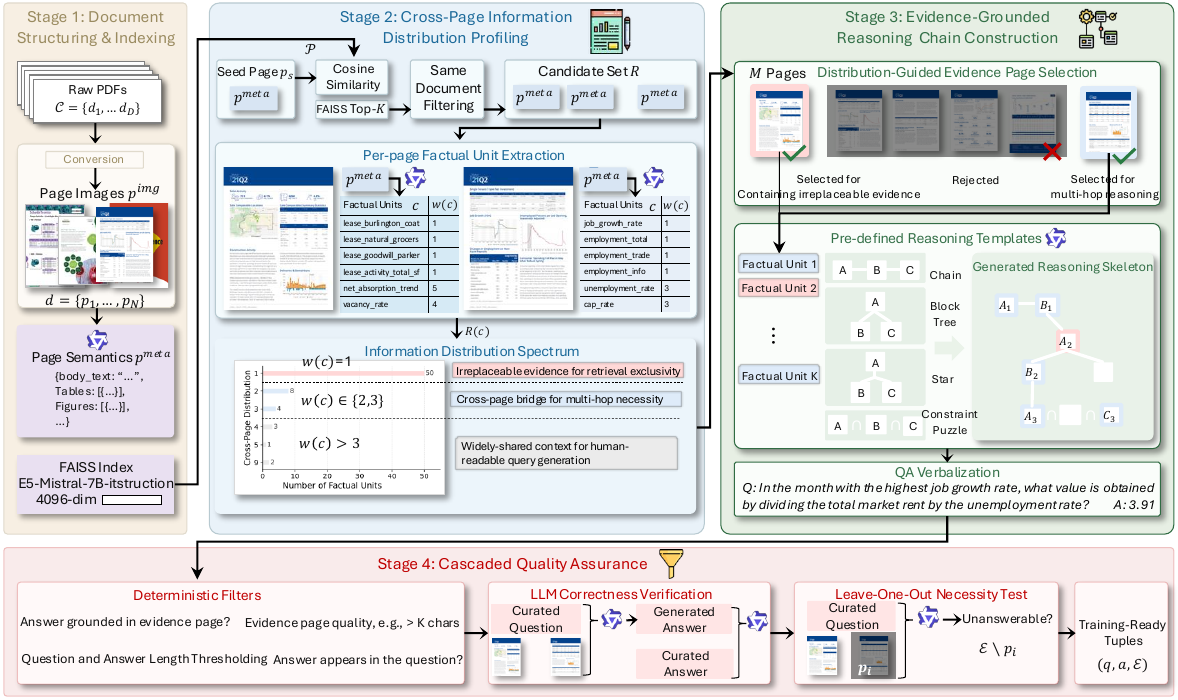} 
\vspace{-5mm}
\caption{ {Overview of the proposed DocArena data curation pipeline. \textbf{Stage~1} converts raw PDFs into page images, MLLM-extracted structured text, and a dense retrieval index. \textbf{Stage~2} profiles cross-page information distribution and identifies irreplaceable evidence ($w{=}1$) to guarantee evidence exclusivity. \textbf{Stage~3} constructs diverse, reasoning-intensive QA pairs grounded in the distribution profile with template-controlled diversity. \textbf{Stage~4} applies three-layer cascaded verification to eliminate noise without human annotation.}
 }
\vspace{-5mm}
\label{fig: data curation} 
\end{figure*}

\textbf{Pipeline Overview.}
As shown in Fig.~\ref{fig: data curation}, the goal of the automated curation pipeline is to turn a raw document collection into a controllable, scalable training environment without any human annotation. Given a collection of raw documents $\mathcal{C} = \{d_1, d_2, \ldots, d_D\}$, the pipeline outputs training samples $\{(q_i, a_i, \mathcal{E}_i)\}$ through four stages (see Fig.~\ref{fig: data curation}): (1) document structuring and indexing, (2) cross-page information distribution profiling, (3) reasoning chain construction, and (4) cascaded quality assurance.

\noindent\textbf{Stage I: Document Structuring and Indexing.}
We perform an offline, one-time preprocessing over the entire source document collection, producing three representations shared by all subsequent stages. (1)~\textit{Page images}---each page is rendered as an image, enabling downstream MLLM perception and interaction with multimodal tools. (2)~\textit{Structured semantic text}---an MLLM extracts body text, table content, figure descriptions, etc., from each page image into a per-page knowledge base. (3)~\textit{A dense retrieval index}---page-level semantic text is encoded by a text retriever into a FAISS inner-product index, serving the pipeline's internal cross-page retrieval needs.

\noindent\textbf{Stage II: Cross-Page Information Distribution Profiling.}
Building $(q, a, \mathcal{E})$ triplets from raw documents is non-trivial. The first challenge specific to search agents is to ensure \textit{evidence exclusivity}: the answer can only be derived from the annotated evidence pages $\mathcal{E}$, not from other pages in the same document, to avoid data noise and potential reward hacking. A valid training sample needs to satisfy the following condition
\begin{equation}\label{eq: exclusivity}
I(a;\, \mathcal{P} \setminus \mathcal{E} \mid q) \approx 0,
\end{equation}
meaning that the answer carries near-zero mutual information with non-evidence pages given the question. The proposed curation pipeline achieves this by profiling the information distribution across pages \textit{before} construction.

Specifically, we first sample a seed page $p_s$ and retrieve its top-$K$ semantically related pages from the same document via the FAISS index, forming a candidate set $R$ ($K{=}15{\sim}30$). We then independently extract \textit{factual units} from each page in $R$ using an MLLM. Each factual unit is a minimal, self-contained knowledge point with a standardized \texttt{key} that categorizes the factual unit to the concept level. We then aggregate all factual units by \texttt{key} and compute each concept's \textit{distribution width}
\begin{equation}\label{eq: width}
w(c) = |\mathcal{P}(c)|, \quad \text{where } \mathcal{P}(c) = \{p \in R : c \text{ appears in } p\}.
\end{equation}

The distribution width $w(c)$ partitions all concepts into three bands. Concepts with $w(c)=1$ appear on exactly one page and serve as \textit{irreplaceable evidence}, meaning the page must be uniquely retrieved to complete the reasoning chain. Concepts with $w(c)\in\{2,3\}$ appear on a small number of pages with different values, creating the cross-page links that make multi-hop reasoning possible. Concepts with $w(c)>3$ appear broadly across many pages. They provide natural-language context that can be referred to by the MLLM to make the generated question readable.

\noindent \textbf{Stage III: Evidence-Grounded Reasoning Chain Construction.}
Although the training data only contains (question, answer, evidence) tuples without intermediate trajectories, the quality of these tuples determines whether the agent can learn meaningful search and reasoning behaviors. To construct challenging, reasoning-intensive QA pairs, the proposed curation pipeline provides a principled way to assist the MLLM in building scalable reasoning chains during data construction.

Given the distribution profile from Stage~II, we use a sequence of MLLM calls to jointly select evidence pages, construct a reasoning chain, and verbalize the final QA pair. The evidence page selection is constrained by the distribution profile: at least one piece of irreplaceable evidence ($w{=}1$) must be included, and the selected pages must span at least $M$ distinct pages within the same document. Based on the selected pages and their factual units, the MLLM chooses a reasoning template from several predefined types, such as \texttt{chain}, \texttt{star}, \texttt{block\_tree}, and \texttt{constraint\_puzzle}, and constructs a step-by-step reasoning skeleton. We maintain a global template counter and trigger re-selection when any single template exceeds a predefined threshold, preventing reasoning type collapse. Finally, the skeleton is verbalized into a natural language question and answer. The question embeds information from the irreplaceable evidence to serve as a search clue, but does not reveal page identifiers or contain the answer.

\begin{figure*}[t]
\centering
\includegraphics[width=0.249\textwidth]{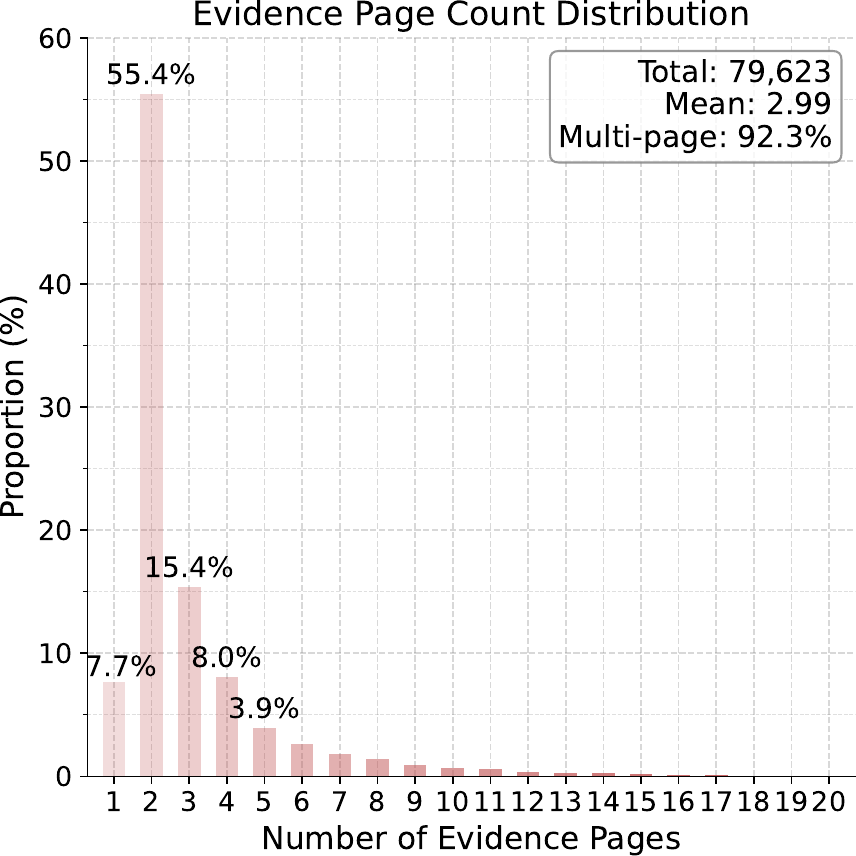}%
\includegraphics[width=0.249\textwidth]{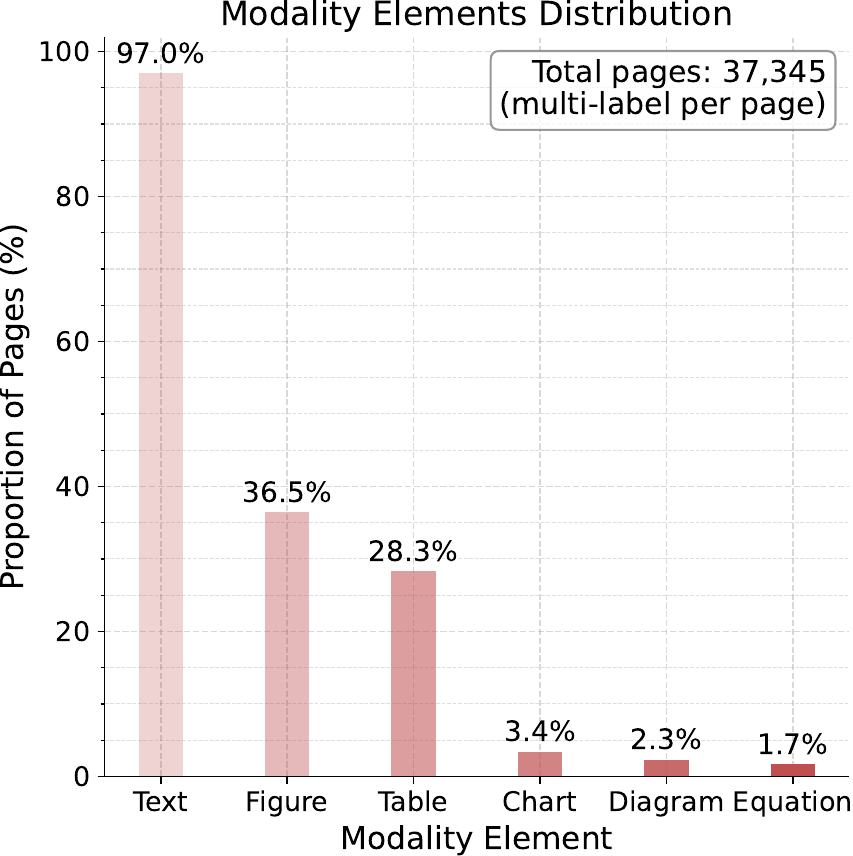}%
\includegraphics[width=0.249\textwidth]{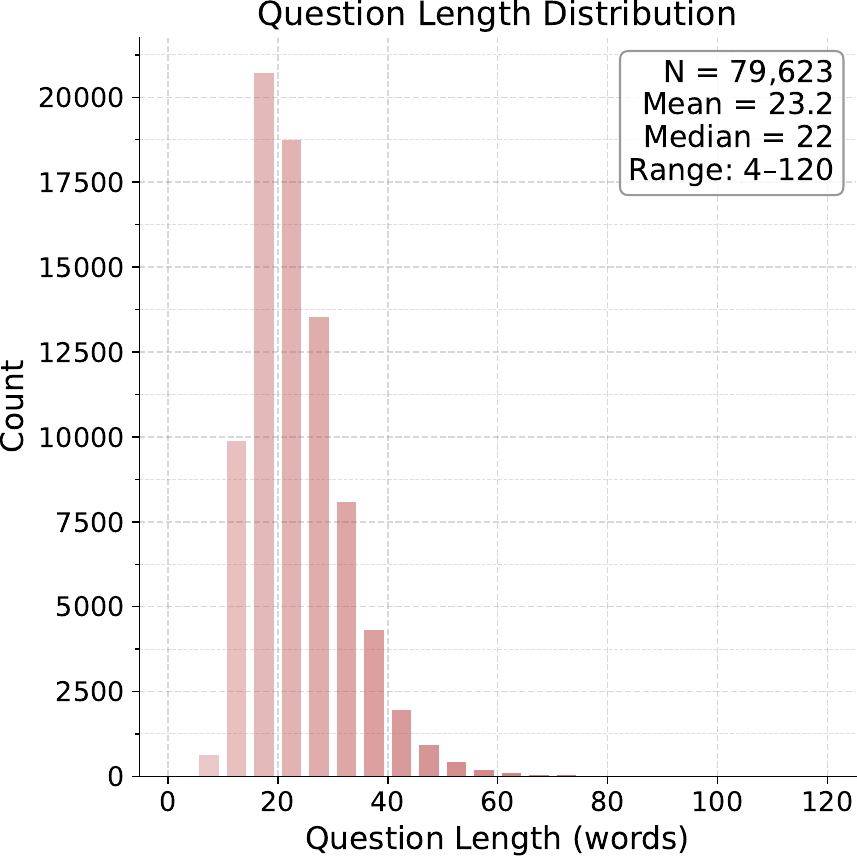}%
\includegraphics[width=0.249\textwidth]{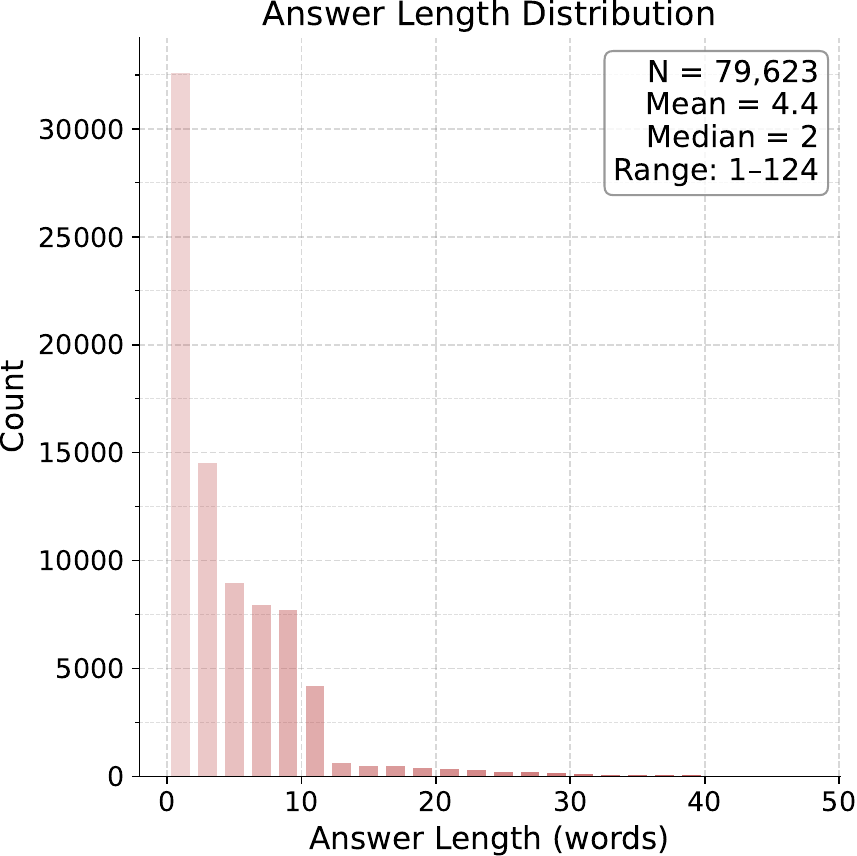}%
\\
\includegraphics[width=0.249\textwidth]{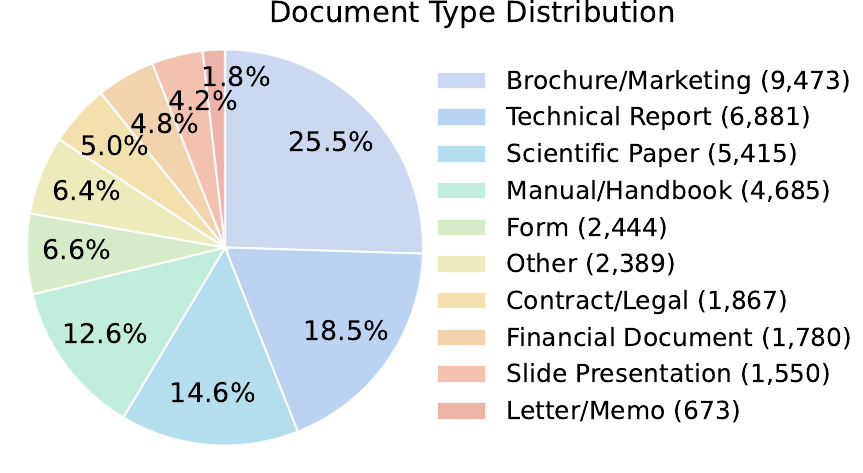}%
\includegraphics[width=0.249\textwidth]{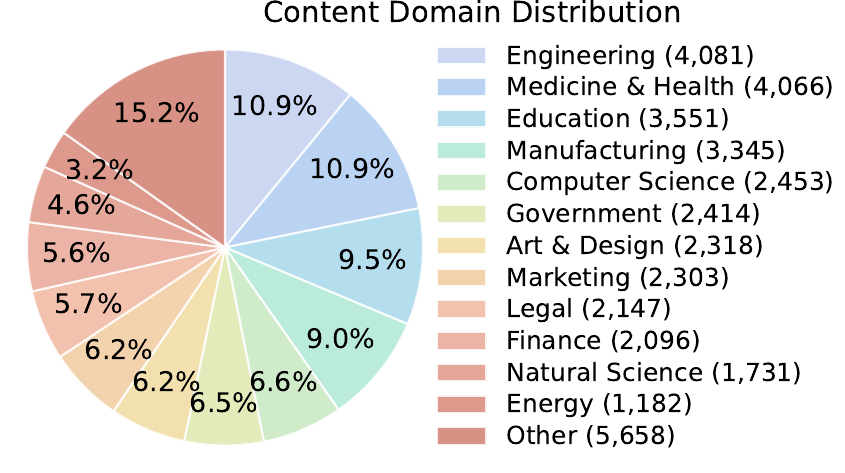}%
\includegraphics[width=0.249\textwidth]{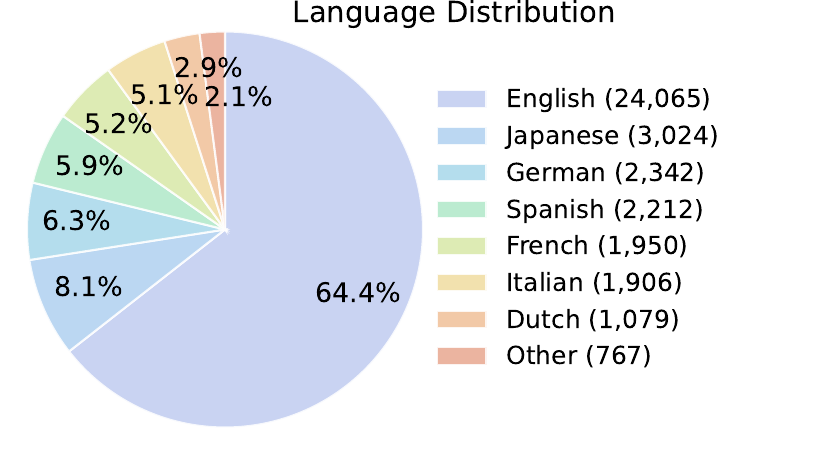}%
\includegraphics[width=0.249\textwidth]{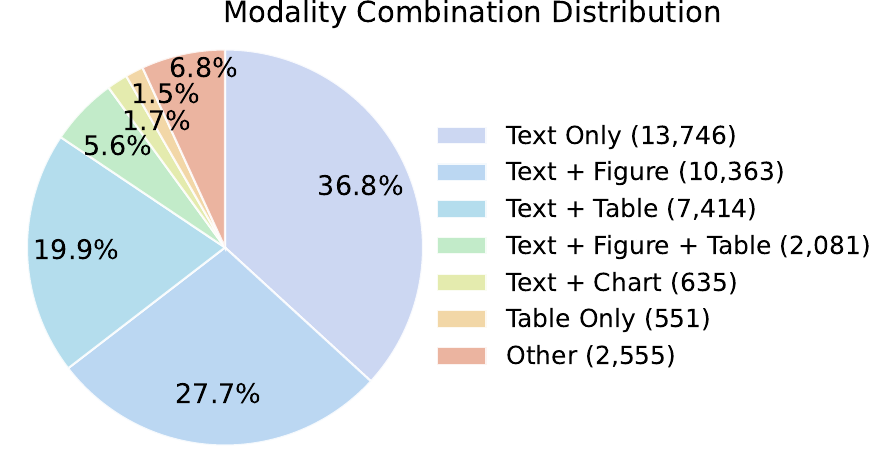}%
\vspace{-2mm}
\caption{Dataset statistics of \textsc{DocArena}. \textbf{Top row}: distributions of evidence pages per question, modality elements per page, question length, and answer length. \textbf{Bottom row}: distributions of document type, content domain, language, and modality combination.}
\label{fig:dataset_statistics}
\vspace{-6mm}
\end{figure*}

\noindent \textbf{Stage IV: Cascaded Quality Assurance.}
DocArena also delivers insights on MLLM-driven data quality assessment, without any human annotation. We design a three-layer  verification system ordered by the increasing computational cost, so that the lightweight checks eliminate low-quality samples early and reduce the load on downstream steps.

\textit{Layer 1: Deterministic filters} are rule-based checks including (a) is the answer grounded in the evidence pages? (b) does the evidence page contain sufficient content (rejecting boilerplate pages with text below a character threshold)? (c) do the question and answer meet length constraints? (d) does the answer already appear in the question?
\textit{Layer 2: MLLM correctness verification.} An MLLM answers the question given the evidence pages, and a separate MLLM call judges whether its response is consistent with the curated answer. Samples where the MLLM cannot reproduce the answer are rejected.
\textit{Layer 3: Leave-one-page-out necessity test.} For each evidence page $p_j \in \mathcal{E}$, we remove it and test whether the question remains answerable from $\mathcal{E} \setminus \{p_j\}$ alone. If any single page can be removed without affecting answerability, it is redundant and the sample is rejected, which ensures every evidence page is truly necessary.

\textbf{Curated Training Dataset.}
We apply the proposed pipeline to CCpdf~\cite{turski2023ccpdf}, a large-scale corpus of visually rich PDF documents. The curated dataset, DocArena-79K, contains $79,623$ QA pairs from $8,336$ source documents, covering $37,480$ unique evidence pages with $238,271$ total page references. Among all samples, $92.3\%$ require multi-page evidence (mean $2.99$ pages per sample, range $1\sim20$). The evidence pages span diverse content, with $63.2\%$ containing at least one table, figure, or chart beyond plain text. The source documents cover $16$ content domains (engineering, medicine, education, finance, etc.) with no single domain exceeding $11\%$, and $10$ document types (technical reports, scientific papers, manuals, forms, etc.) across $49$ languages. We provide the statistics of the DocArena-79K in Figure~\ref{fig:dataset_statistics} and more details in the appendix.

\begin{figure*}[t] 
\centering 
\includegraphics[width=\textwidth]{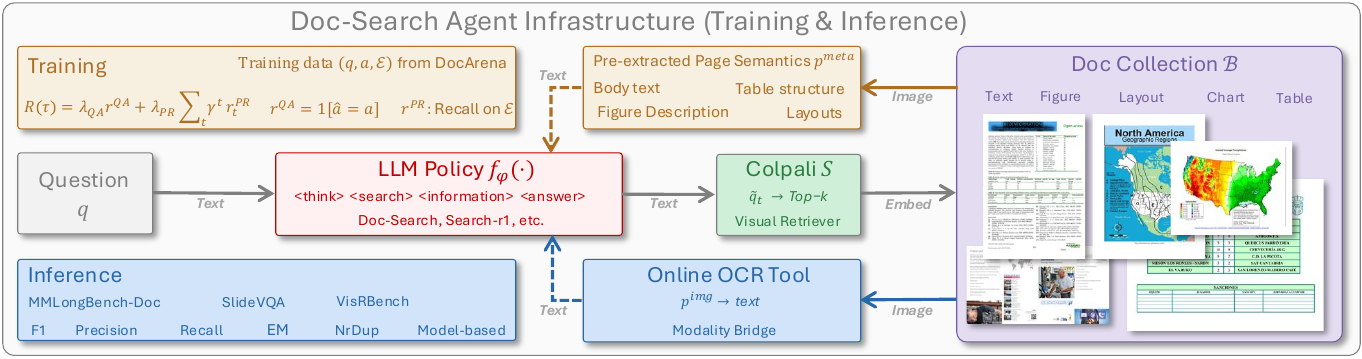} 
\vspace{-5mm}
\caption{Overview of the Doc-Search agent infrastructure. The system addresses multimodal document retrieval and QA tasks. We adopt a multimodal retriever (ColPali), an OCR tool , and a LLM-based policy model for multi-turn interaction, which decouples the visual perception from the policy model and allows different policy model under identical system configurations. During training (top), the policy interacts with the DocArena environment produced offline upon automatic curation pipeline (Section~\ref{subsec: pipeline}). Modality transitions are annotated throughout the system.
}
\vspace{-5mm}
\label{fig: framework} 
\end{figure*}

\subsection{The Doc-Search Agent Infrastructure}\label{subsec: agent}

Beyond the data curation pipeline, we design a Doc-Search agent infrastructure for multimodal document retrieval and QA as shown in Figure~\ref{fig: framework}. The infrastructure decouples visual perception from the policy model, allowing text-based LLM to serve as the reasoning backbone, maintaining compatibility with existing LLM checkpoints for convenient deployment and fair benchmarking.

\textbf{Infrastructure.}
As shown in Figure~\ref{fig: framework}, the infrastructure consists of three components: a text-based LLM policy $f_\phi$, a multimodal retriever ColPali~\cite{faysse2024colpali}, and an online OCR tool. We choose an LLM-based agent for its strong multi-step reasoning ability~\cite{cao2025large} and efficiency over MLLM-based alternatives that require processing high-resolution page images. At each step $t$, the agent generates a search query $\tilde{q}_t$, and ColPali retrieves the most relevant page images from the document collection $\mathcal{B}$. The agent then applies online OCR to extract text from the retrieved pages and continues reasoning with the updated context. This process repeats until the agent produces a final answer or reaches the maximum search turn budget. During training, retrieved content tokens are masked from the policy loss following previous works~\cite{jin2025search}.

\textbf{Reward Design.}
Given that the curated DocArena-79K environment involves multi-turn retrieval and long-horizon reasoning, we adopt a retrieval-based progress reward in addition to the standard QA reward, to better incentivize the agent to actively mine useful information across search turns. We have the reward
\begin{equation}
\label{eq:total_reward}
R(\tau) = \lambda_{\text{QA}} \cdot r^{\text{QA}}(\tau) + \lambda_{\text{PR}} \cdot \sum\nolimits_{t=1}^{T} \gamma^{\,t}\, r^{\text{PR}}_t,
\end{equation}
where $r^{\text{QA}} = \mathbb{1}[\hat{a}=a]$ is the exact-match QA reward, $r^{\text{PR}}_t$ is the recall-based retrieval reward at search turn $t$, and $\gamma$ is a discount factor. Duplicate queries receive zero retrieval reward. The progress reward is assigned to the search action token for fine-grained credit assignment. We provide more details in the appendix.

In summary, this work addresses the training data challenge for search agents at multiple levels. We define a five-dimensional quality model that characterizes what makes a good training environment (Section~\ref{subsec: env}), design a fully automated curation pipeline that constructs and controls such environments from raw documents (Section~\ref{subsec: pipeline}), and provide DocArena-79K as a ready-to-use training environment. The effectiveness of the constructed environment is validated through the search behaviors of agents trained upon it (Section~\ref{sec:experiment}). Additional design insights and implementation details are provided in the appendix.

\section{Experiment}\label{sec:experiment}

\subsection{Experimental Settings}\label{subsec: settings}

\textbf{Evaluation Benchmarks.}
We evaluate the search agents on six multimodal document scenarios and seven text-based QA benchmarks.
For multimodal document evaluation, we adopt MMLongBench-Doc~\cite{ma2024mmlongbench}, VisRBench~\cite{chen2025visr}, and SlideVQA~\cite{tanaka2023slidevqa}.
VisRBench contains $53$K QA pairs across $\sim1,286$ documents; we evaluate on three sub-scenarios: FigureQA, TableQA, and TextQA, each requiring different types of visual-structural reasoning.
MMLongBench-Doc contains $1,082$ expert-annotated questions with both single-page (SP) and multi-page (MP) scenarios.
SlideVQA provides $2,215$ multi-page questions over slide decks.
For text-based QA, we adopt seven benchmarks following previous works~\cite{jin2025search}, \emph{e.g.}, Natural Questions~\cite{kwiatkowski2019natural}, TriviaQA~\cite{joshi2017triviaqa}, PopQA~\cite{mallen2023not}, HotpotQA~\cite{yang2018hotpotqa}, 2WikiMultiHopQA~\cite{ho2020constructing}, MuSiQue~\cite{trivedi2022musique}, and Bamboogle~\cite{press2023measuring}, covering both single-hop and multi-hop reasoning.

\begin{table*}[t]
\caption{
Performance comparison on \textbf{MMLongBench-Doc} for both Single-page (SP) and Multi-page (MP) scenarios. All methods share the same inference infrastructure and differ only in the policy model. Best results are \textbf{bolded} and second best are \underline{underlined}.
}
\vspace{-2mm}
\resizebox{\textwidth}{!}{
\begin{tabular}{c|ccc|ccc|cccc|ccc}
\hline
\multirow{3}{*}{Method}
& \multicolumn{6}{c|}{Single-page}
& \multicolumn{7}{c}{Multi-page} \\
\cline{2-14}
& \multicolumn{3}{c|}{Retrieval}
& \multicolumn{3}{c|}{QA Metrics}
& \multicolumn{4}{c|}{Retrieval}
& \multicolumn{3}{c}{QA Metrics} \\
\cline{2-14}
& Top-1 $\uparrow$ & Top-5 $\uparrow$ & NrDup $\downarrow$
& EM $\uparrow$ & Model $\uparrow$ & PNLS $\uparrow$
& Recall $\uparrow$ & Precision $\uparrow$ & F1 $\uparrow$ & NrDup $\downarrow$
& EM $\uparrow$ & Model $\uparrow$ & PNLS $\uparrow$ \\
\hline
Search-R1~\cite{jin2025search}
& 69.82 & 79.67 & 14.68 & 25.94 & 23.63 & 34.81
& 55.24 & 32.18 & 37.09 & 22.77 & 20.10 & 17.92 & 28.98 \\
DeepResearcher~\cite{zheng2025deepresearcher}
& 69.61 & 81.93 & 38.84 & 25.50 & 20.81 & 33.23
& 55.82 & 33.37 & 38.25 & 44.39 & 19.71 & 16.25 & 27.16 \\
IKEA~\cite{huang2025reinforced}
& 69.82 & 81.52 & 19.85 & \underline{28.30} & \textbf{24.64} & \underline{37.26}
& 54.43 & \textbf{34.27} & \underline{38.41} & 23.62 & \underline{21.03} & \underline{18.89} & 30.02 \\
R-Search~\cite{zhao2025r}
& 63.45 & 75.15 & 21.28 & 26.02 & 21.50 & 35.25
& 49.58 & 30.79 & 34.84 & 30.33 & 18.90 & 16.02 & 28.56 \\
ZeroSearch~\cite{sun2025zerosearch}
& 62.63 & 76.18 & 31.03 & 24.37 & 21.05 & 33.44
& 48.49 & 31.33 & 34.74 & 26.32 & 18.73 & 15.89 & 28.14 \\
AutoRefine~\cite{shi2025search}
& 65.71 & 76.80 & 14.81 & 26.60 & 22.29 & 35.84
& 52.23 & 32.66 & 36.75 & 15.04 & 20.47 & 17.50 & 29.82 \\
StepSearch~\cite{wang2025stepsearch}
& 63.86 & 77.00 & \underline{3.82} & 25.13 & 21.17 & 35.09
& 52.09 & 30.31 & 35.24 & \underline{5.30} & 20.96 & 17.30 & \underline{30.50} \\
ASearcher~\cite{gao2025beyond}
& 56.06 & 63.24 & 41.99 & 15.88 & 12.76 & 22.36
& 45.38 & 25.52 & 29.87 & 46.78 & 14.46 & 12.18 & 21.30 \\
ReSearch~\cite{chen2025learning}
& \underline{70.64} & \underline{82.96} & 31.25 & 22.86 & 19.11 & 30.85
& \underline{56.53} & 33.45 & 38.22 & 45.30 & 18.41 & 15.80 & 26.22 \\
Ours
& \textbf{75.77} & \textbf{85.22} & \textbf{2.46} & \textbf{29.34} & \underline{24.27} & \textbf{38.27}
& \textbf{61.38} & \underline{33.76} & \textbf{39.68} & \textbf{2.97} & \textbf{24.18} & \textbf{19.43} & \textbf{33.12} \\
\hline
\end{tabular}
}
\vspace{-3mm}
\label{tab:mmlongbench_combined}
\end{table*}

\begin{table*}[t]
\caption{
Performance comparison on \textbf{VisRBench} FigureQA and TableQA (both single-page). All methods share the same inference infrastructure and differ only in the policy model.
}
\vspace{-2mm}
\resizebox{\textwidth}{!}{
\begin{tabular}{c|ccc|ccc|ccc|ccc}
\hline
\multirow{3}{*}{Method}
& \multicolumn{6}{c|}{Figure QA}
& \multicolumn{6}{c}{Table QA} \\
\cline{2-13}
& \multicolumn{3}{c|}{Retrieval}
& \multicolumn{3}{c|}{QA Metrics}
& \multicolumn{3}{c|}{Retrieval}
& \multicolumn{3}{c}{QA Metrics} \\
\cline{2-13}
& Top-1 $\uparrow$ & Top-5 $\uparrow$ & NrDup $\downarrow$
& EM $\uparrow$ & Model $\uparrow$ & PNLS $\uparrow$
& Top-1 $\uparrow$ & Top-5 $\uparrow$ & NrDup $\downarrow$
& EM $\uparrow$ & Model $\uparrow$ & PNLS $\uparrow$ \\
\hline
Search-R1~\cite{jin2025search}
& 77.70 & 90.41 & 12.95 & 6.04 & 11.24 & 12.36
& \underline{54.25} & 74.81 & 14.85 & 30.05 & 31.17 & 36.17 \\
DeepResearcher~\cite{zheng2025deepresearcher}
& 78.18 & 91.13 & 27.59 & 9.79 & 11.10 & \underline{18.09}
& 52.47 & 73.03 & 35.00 & \underline{33.44} & 31.75 & 38.85 \\
IKEA~\cite{huang2025reinforced}
& 75.54 & 91.61 & 8.40 & 7.69 & 13.72 & 14.09
& 54.11 & 74.13 & 19.65 & 31.74 & \underline{33.14} & 37.91 \\
R-Search~\cite{zhao2025r}
& 73.38 & 84.65 & 18.12 & \underline{11.02} & \underline{14.42} & 16.24
& 47.40 & 68.29 & 28.71 & 31.24 & 30.66 & 37.00 \\
ZeroSearch~\cite{sun2025zerosearch}
& 73.38 & 86.81 & 18.52 & 10.59 & 13.90 & 16.52
& 45.36 & 65.89 & 18.66 & 31.82 & 30.96 & 38.07 \\
AutoRefine~\cite{shi2025search}
& 75.30 & 88.25 & 7.74 & 8.38 & 11.83 & 14.79
& 50.20 & 71.23 & 15.37 & 32.47 & 32.20 & \underline{39.30} \\
StepSearch~\cite{wang2025stepsearch}
& 77.22 & 90.65 & \textbf{2.35} & 10.62 & 12.74 & 16.32
& 30.44 & 43.78 & \underline{8.25} & 25.22 & 24.75 & 31.18 \\
ASearcher~\cite{gao2025beyond}
& 58.03 & 64.75 & 29.79 & 11.00 & 13.06 & \textbf{18.25}
& 46.64 & 61.08 & 52.46 & 26.21 & 25.43 & 32.80 \\
ReSearch~\cite{chen2025learning}
& \underline{79.62} & \underline{96.40} & 23.56 & 8.38 & 11.07 & 16.15
& 53.98 & \underline{75.95} & 25.22 & 32.19 & 31.72 & 38.34 \\
Ours
& \textbf{86.09} & \textbf{97.60} & \underline{2.84} & \textbf{13.95} & \textbf{14.95} & 17.50
& \textbf{64.38} & \textbf{82.73} & \textbf{3.43} & \textbf{35.99} & \textbf{34.40} & \textbf{40.78} \\
\hline
\end{tabular}
}
\vspace{-2mm}
\label{tab:visrbench_combined}
\end{table*}

\begin{table*}[t]
\caption{
Performance comparison on \textbf{VisRBench} TextQA (single-page) and \textbf{SlideVQA} (multi-page). All methods share the same inference infrastructure and differ only in the policy model.
}
\vspace{-2mm}
\resizebox{\textwidth}{!}{
\begin{tabular}{c|ccc|ccc|cccc|ccc}
\hline
\multirow{3}{*}{Method}
& \multicolumn{6}{c|}{VisRBench TextQA (SP)}
& \multicolumn{7}{c}{SlideVQA (MP)} \\
\cline{2-14}
& \multicolumn{3}{c|}{Retrieval}
& \multicolumn{3}{c|}{QA Metrics}
& \multicolumn{4}{c|}{Retrieval}
& \multicolumn{3}{c}{QA Metrics} \\
\cline{2-14}
& Top-1 $\uparrow$ & Top-5 $\uparrow$ & NrDup $\downarrow$
& EM $\uparrow$ & Model $\uparrow$ & PNLS $\uparrow$
& Recall $\uparrow$ & Precision $\uparrow$ & F1 $\uparrow$ & NrDup $\downarrow$
& EM $\uparrow$ & Model $\uparrow$ & PNLS $\uparrow$ \\
\hline
Search-R1~\cite{jin2025search}
& {74.34} & {91.44} & 9.99 & 43.83 & 44.40 & 47.68
& 81.81 & 53.85 & 61.04 & 16.01 & 43.55 & 40.12 & 49.62 \\
DeepResearcher~\cite{zheng2025deepresearcher}
& 70.68 & 87.09 & 27.56 & 48.21 & 44.73 & 50.15
& 78.17 & 52.11 & 58.91 & 36.63 & 41.32 & 36.79 & 45.75 \\
IKEA~\cite{huang2025reinforced}
& 73.73 & 89.83 & 10.64 & 44.27 & 45.84 & 48.24
& 76.44 & 52.62 & 58.68 & 9.55 & 43.01 & 39.84 & 48.48 \\
R-Search~\cite{zhao2025r}
& 68.77 & 86.72 & 20.94 & 46.91 & 43.56 & 48.55
& 74.54 & 51.20 & 57.04 & 16.73 & 44.26 & 39.44 & 48.49 \\
ZeroSearch~\cite{sun2025zerosearch}
& 56.01 & 71.42 & 34.44 & 43.70 & 40.98 & 48.05
& 64.00 & 45.17 & 49.67 & \underline{8.97} & 40.81 & 36.67 & 47.14 \\
AutoRefine~\cite{shi2025search}
& 70.40 & 87.26 & 11.09 & 47.58 & 45.82 & \underline{51.39}
& 75.16 & 50.23 & 56.68 & 9.59 & 42.84 & 38.43 & 47.73 \\
StepSearch~\cite{wang2025stepsearch}
& 68.34 & 87.88 & \textbf{0.61} & \textbf{51.16} & \underline{47.65} & \textbf{53.83}
& 80.82 & 52.49 & 60.01 & \textbf{2.79} & \underline{47.90} & \textbf{42.64} & \underline{52.11} \\
ASearcher~\cite{gao2025beyond}
& 66.02 & 75.92 & 32.06 & 36.00 & 34.54 & 40.92
& 63.82 & 41.63 & 47.29 & 27.66 & 35.41 & 31.18 & 40.31 \\
ReSearch~\cite{chen2025learning}
& \underline{75.55} & \underline{91.95} & 22.04 & 47.61 & 45.45 & 50.42
& \underline{84.98} & \underline{54.71} & \underline{62.72} & 29.77 & 43.78 & 38.97 & 47.91 \\
Ours
& \textbf{83.27} & \textbf{94.98} & \underline{2.06} & \underline{49.78} & \textbf{47.93} & 49.80
& \textbf{86.28} & \textbf{59.78} & \textbf{66.95} & 21.21 & \textbf{47.96} & \underline{42.43} & \textbf{52.18} \\
\hline
\end{tabular}
}
\vspace{-1mm}
\label{tab:textvqa_slidevqa_combined}
\end{table*}


\begin{table*}[t]
\caption{
Performance comparison on \textbf{text-based QA benchmarks}. All methods use E5-base-v2 with Wikipedia-18 as the retriever, identical inference parameters, and differ only in the policy model. Our method has never seen any text QA training data.
}
\centering
\vspace{-2mm}
\resizebox{\textwidth}{!}{
\begin{tabular}{l|ccccccc|c}
\hline
Method
& Natural Questions $\uparrow$ & TriviaQA $\uparrow$ & PopQA $\uparrow$ & HotpotQA $\uparrow$ & 2WikiMultiHopQA $\uparrow$ & MuSiQue $\uparrow$ & Bamboogle $\uparrow$ & Avg $\uparrow$ \\
\hline
Search-R1~\cite{jin2025search}
& \textbf{39.20} & \underline{59.22} & 38.91 & 36.16 & 27.58 & 14.94 & 36.29 & 36.04 \\
StepSearch~\cite{wang2025stepsearch}
& 36.84 & 57.51 & 36.41 & \underline{36.31} & 32.05 & \textbf{19.53} & 35.48 & 36.30 \\
ZeroSearch~\cite{sun2025zerosearch}
& 36.29 & 57.09 & 35.79 & 32.11 & \underline{34.90} & 10.88 & 29.03 & 33.73 \\
R-Search~\cite{zhao2025r}
& 34.71 & 58.85 & 36.17 & 32.59 & 29.76 & 12.83 & 35.48 & 34.34 \\
ReSearch~\cite{chen2025learning}
& 36.40 & 58.71 & \underline{38.87} & 34.88 & 27.97 & \underline{17.09} & \textbf{40.32} & \underline{36.32} \\
DeepResearcher~\cite{zheng2025deepresearcher}
& 35.60 & 58.44 & 37.06 & 34.85 & 31.39 & 14.40 & \underline{39.52} & 35.89 \\
\hline
Ours
& \underline{36.68} & \textbf{59.63} & \textbf{40.34} & \textbf{37.23} & \textbf{36.82} & 14.11 & 33.87 & \textbf{37.24} \\
\hline
\end{tabular}
}
\label{tab:flashrag_comparison}
\end{table*}

\begin{table}[t]
\centering
\caption{Ablation studies on (a) reward components and (b) PR reward ratio (MMLongBench-Doc MP). PR = progress-based retrieval reward, QA = outcome-based answer quality reward.}
\vspace{-5mm}
\begin{minipage}[t]{0.48\columnwidth}
\centering
\subcaption{Reward modeling}\label{tab:ablation_reward}
\vspace{-3mm}
\resizebox{\textwidth}{!}{
\begin{tabular}{l|cccc|ccc}
\hline
\multirow{2}{*}{Reward} & \multicolumn{4}{c|}{Retrieval} & \multicolumn{3}{c}{QA} \\
\cline{2-8}
& Rec & Prec & F1 & NrDup$\downarrow$ & EM & Mod & PNLS \\
\hline
w/o PR & 53.68 & \textbf{34.19} & 38.49 & 36.36 & 22.26 & 19.14 & 32.68 \\
w/o QA & \textbf{62.95} & 30.74 & 37.19 & 5.57 & 5.88 & 4.49 & 12.04 \\
PR+QA & 61.38 & 33.76 & \textbf{39.68} & \textbf{2.97} & \textbf{24.18} & \textbf{19.43} & \textbf{33.12} \\
\hline
\end{tabular}
}
\end{minipage}
\hfill
\begin{minipage}[t]{0.48\columnwidth}
\centering
\subcaption{PR reward ratio}\label{tab:ablation_pr_ratio}
\vspace{-3mm}
\resizebox{\textwidth}{!}{
\begin{tabular}{l|cccc|ccc}
\hline
\multirow{2}{*}{PR Ratio} & \multicolumn{4}{c|}{Retrieval} & \multicolumn{3}{c}{QA} \\
\cline{2-8}
& Rec & Prec & F1 & NrDup$\downarrow$ & EM & Mod & PNLS \\
\hline
0.1 & \textbf{61.93} & \textbf{33.87} & \textbf{39.81} & 11.07 & 21.50 & 18.03 & 30.67 \\
0.3 & 60.70 & 33.00 & 39.15 & 7.24 & 21.85 & 18.27 & 31.05 \\
0.5 & 61.38 & 33.76 & 39.68 & \textbf{2.97} & \textbf{24.18} & \textbf{19.43} & \textbf{33.12} \\
\hline
\end{tabular}
}
\end{minipage}
\label{tab:ablation_row1}
\end{table}

\begin{table}[t]
\centering
\caption{Ablation studies on (a) optimizer and (b) retriever backbone (MMLongBench-Doc MP).}
\vspace{-6mm}
\begin{minipage}[t]{0.48\columnwidth}
\centering
\subcaption{Optimizer}\label{tab:ablation_optimizer}
\vspace{-3mm}
\resizebox{\textwidth}{!}{
\begin{tabular}{l|cccc|ccc}
\hline
\multirow{2}{*}{Optimizer} & \multicolumn{4}{c|}{Retrieval} & \multicolumn{3}{c}{QA} \\
\cline{2-8}
& Rec & Prec & F1 & NrDup$\downarrow$ & EM & Mod & PNLS \\
\hline
PPO & 61.22 & \textbf{34.01} & 39.54 & \textbf{2.16} & 23.81 & \textbf{19.51} & 32.53 \\
GRPO & \textbf{61.38} & 33.76 & \textbf{39.68} & 2.97 & \textbf{24.18} & 19.43 & \textbf{33.12} \\
\hline
\end{tabular}
}
\end{minipage}
\hfill
\begin{minipage}[t]{0.48\columnwidth}
\centering
\subcaption{Retriever}\label{tab:ablation_retriever}
\vspace{-3mm}
\resizebox{\textwidth}{!}{
\begin{tabular}{l|cccc|ccc}
\hline
\multirow{2}{*}{Retriever} & \multicolumn{4}{c|}{Retrieval} & \multicolumn{3}{c}{QA} \\
\cline{2-8}
& Rec & Prec & F1 & NrDup$\downarrow$ & EM & Mod & PNLS \\
\hline
E5-Text & 58.24 & 32.77 & 38.25 & \textbf{1.40} & 21.80 & 17.99 & 31.32 \\
ColPali & \textbf{61.38} & \textbf{33.76} & \textbf{39.68} & 2.97 & \textbf{24.18} & \textbf{19.43} & \textbf{33.12} \\
\hline
\end{tabular}
}
\end{minipage}
\vspace{-4mm}
\label{tab:ablation_row2}
\end{table}

\textbf{Metrics.}
For single-page retrieval, we report Top-$K$ accuracy (\emph{e.g.}, Top-1 and Top-5), measuring whether the evidence page appears among the top-$k$ retrieved pages. For multi-page retrieval, we compute recall, precision, and $F_1$ over the set of evidence pages. For QA quality, we adopt Exact Match (EM)~\cite{yu2024rankrag}, Partial Normalized Levenshtein Similarity (PNLS)~\cite{chen2024mmr}, and a model-based score~\cite{wang2025vrag}. We additionally report the Near-Duplicate Rate (NrDup$\downarrow$), which measures the percentage of multi-round queries where at least one pair has token-level Jaccard similarity $>0.8$. For single-page scenarios, NrDup reflects whether the agent generates repetitive queries across search turns; for multi-page scenarios, it further indicates whether the agent can formulate diverse queries to cover different evidence pages.

\textbf{Compared Methods.}
We compare with nine RL-trained search agents: Search-R1~\cite{jin2025search}, DeepResearcher~\cite{zheng2025deepresearcher}, IKEA~\cite{huang2025reinforced}, R-Search~\cite{zhao2025r}, ZeroSearch~\cite{sun2025zerosearch}, AutoRefine~\cite{shi2025search}, StepSearch~\cite{wang2025stepsearch}, ASearcher~\cite{gao2025beyond}, and ReSearch~\cite{chen2025learning}. All methods are evaluated under the unified Doc-Search agent infrastructure (Section~\ref{subsec: agent}), sharing the same multimodal retriever (ColPali), OCR tool, and inference configurations, with the policy model as the sole controlled variable. For all methods, we adopt publicly released checkpoints trained on Qwen2.5-7B-Instruct\footnote{Qwen2.5-7B is adopted when the instructional checkpoint is not available.}~\cite{team2024qwen2}.

\textbf{Implementation.}
We adopt Qwen2.5-7B-Instruct as the base policy model. We train with Group Relative Policy Optimization (GRPO) using a batch size of $512$, group size of $5$, and learning rate $1\times10^{-6}$. The maximum number of search turns during training is $2$. The retriever returns top-$3$ relevant pages per query. The reward weights are set to $\lambda_{\text{QA}}=0.3$ and $\lambda_{\text{PR}}=0.7$ with a progress reward weight of $0.5$ (Eq.~\ref{eq:total_reward}). During inference, the maximum search turns is $2$ for SP scenarios and $3$ for MP scenarios. For text-based QA evaluation, we adopt E5-base-v2~\cite{wang2022text} with a Wikipedia corpus. Our implementation is built on veRL~\cite{sheng2024hybridflow}. Training is performed on $8\times$A100 GPUs.

\subsection{Main Results}\label{subsec: performance}

\textbf{Multimodal Document Scenarios.}
Tables~\ref{tab:mmlongbench_combined}$\sim$\ref{tab:textvqa_slidevqa_combined} compare search agents across six multimodal document scenarios. Overall, the proposed Doc-Search agent achieves the best retrieval performances, \emph{e.g.}, on MMLongBench-Doc MP, Doc-Search obtains Recall $61.38$ and EM $24.18$, outperforming ReSearch by $+4.85$ on Recall and IKEA by $+3.15$ on EM. The QA metrics follow consistent trends across all scenarios.
Notably, our method achieves $2.06$--$3.43$ on NrDup across all scenarios, while other baselines range from $14$ to $45$, indicating that agents trained on DocArena-79K learn diverse search strategies rather than rigid querying patterns. An interesting outlier is StepSearch, which achieves the highest QA on TextQA (EM $51.16$) yet suffers  degradation on TableQA, suggesting it might be biased toward text-dominant reasoning. By contrast, the proposed search agent maintains competitive performance across different scenarios.

Besides, we observe that the proposed Doc-Search agent consistently achieves NrDup below $3.5\%$ across all six multimodal document scenarios, while the average NrDup of compared baselines can range from $10\%$ to $40\%$. This consistency across benchmarks with distinct evidence structures (single-page figures, multi-page documents, slide decks) suggests that the low query repetition is a stable property of the learned search policy rather than a benchmark-specific artifact, reinforcing the connection between training data quality and agent search behavior.

\textbf{Text-Based QA Generalization.}
Table~\ref{tab:flashrag_comparison} compares the proposed methods with text-based search agents  on seven text-based QA benchmarks. Notably, Doc-Search agent is trained exclusively on DocArena-79K without seeing  text QA training data and achieves the highest average EM ($37.24$), ranking first on four out of seven benchmarks (TriviaQA, PopQA, HotpotQA, 2WikiMultiHopQA). On the multi-hop benchmark 2WikiMultiHopQA, our method reaches $36.82$, outperforming ZeroSearch ($34.90$) by $+1.92$ and Search-R1 ($27.58$) by $+9.24$.
The cross-domain generalization demonstrates that the search strategies learned from a well-constructed multimodal document environment can transfer to unseen text-based scenarios.

\subsection{Model Analysis}\label{subsec: analysis}

\textbf{Reward Modeling Dicussion.}
Table~\ref{tab:ablation_reward} discusses the different reward components on MMLongBench-Doc MP. Removing the retrieval progress reward (w/o PR) causes NrDup to surge from $2.97\%$ to $36.36\%$, meaning the agent degenerates into repetitive querying and loses the ability to formulate diverse searches across turns. This  demonstrates that the PR reward facilitates learning adaptive multi-turn search behaviors. 
The low NrDup of our full model thus indirectly validates the evidence exclusivity guaranteed by the distribution profiling stage of the curation pipeline (Section~\ref{subsec: pipeline}).
Conversely, removing the QA reward (w/o QA) yields the highest Recall but collapses QA performance, as the agent learns to retrieve aggressively without grounding its reasoning in the answer. Combining both rewards achieves the best F1 and QA metrics, confirming that retrieval and QA supervision are complementary.

\textbf{PR Reward Ratio.}
Table~\ref{tab:ablation_pr_ratio} varies the PR reward ratio. Increasing the ratio from $0.1$ to $0.5$ steadily reduces NrDup ($11.07\% \to 7.24\% \to 2.97\%$) and improves QA performance (EM $21.50 \to 21.85 \to 24.18$), with the best overall performance at $0.5$. This suggests that when the training data provides clean evidence annotations, the agent can benefit from stronger retrieval supervision without collapsing, a property enabled by the evidence exclusivity of the DocArena environment.

\textbf{Optimizer and Retriever.}
Table~\ref{tab:ablation_optimizer} compares GRPO and PPO, showing comparable performance with GRPO achieving slightly higher EM ($24.18$ vs.\ $23.81$). Table~\ref{tab:ablation_retriever} compares ColPali (visual) and E5 (text) as the retriever backbone. ColPali outperforms E5 on all metrics (Recall $61.38$ vs.\ $58.24$, EM $24.18$ vs.\ $21.80$), confirming that visual retrieval better captures the semantics of document pages with multimodal data.

\begin{figure*}[t]
    \centering
    \begin{minipage}[t]{0.49\textwidth}
        \centering
        \includegraphics[width=\textwidth]{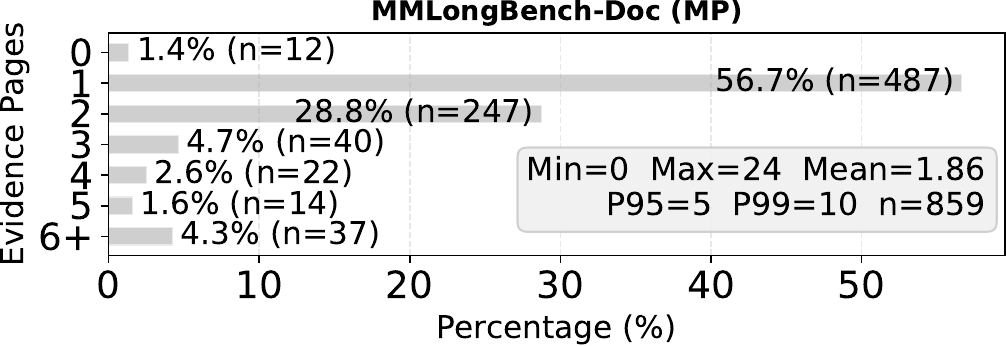}

        \vspace{0.5mm}

        \begin{minipage}[t]{0.49\textwidth}
            \centering
            \includegraphics[width=\textwidth]{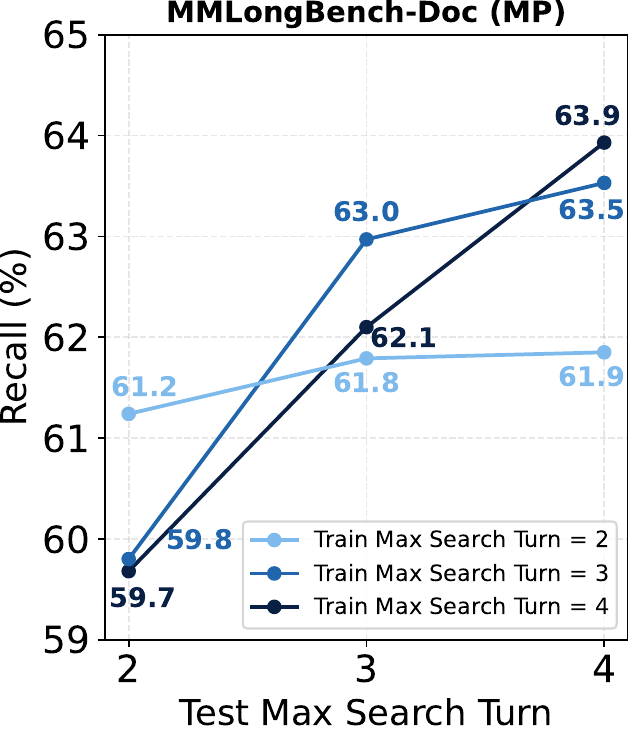}
        \end{minipage}
        \hfill
        \begin{minipage}[t]{0.49\textwidth}
            \centering
            \includegraphics[width=\textwidth]{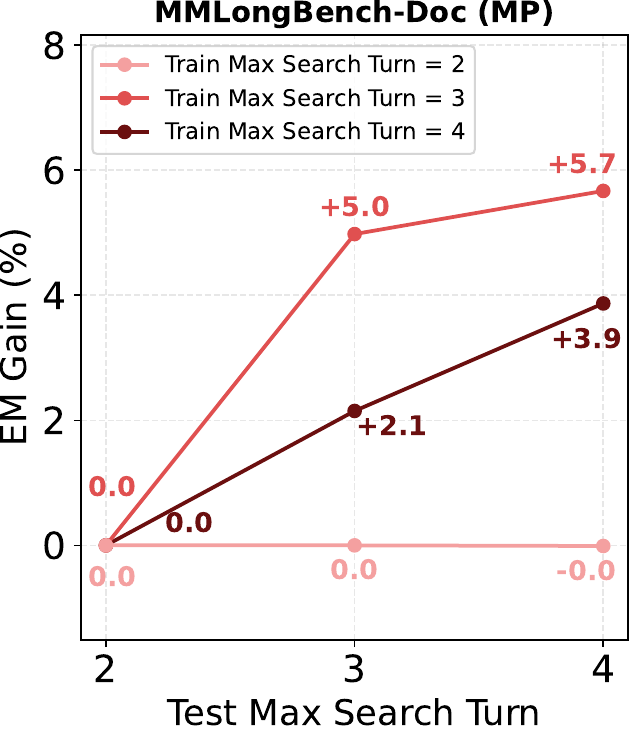}
        \end{minipage}
    \end{minipage}
    \hfill
    \begin{minipage}[t]{0.49\textwidth}
        \centering
        \includegraphics[width=\textwidth]{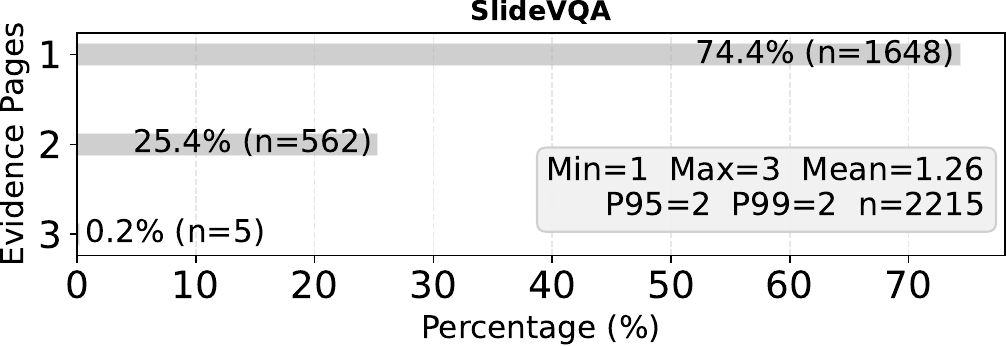}

        \vspace{0.5mm}

        \begin{minipage}[t]{0.49\textwidth}
            \centering
            \includegraphics[width=\textwidth]{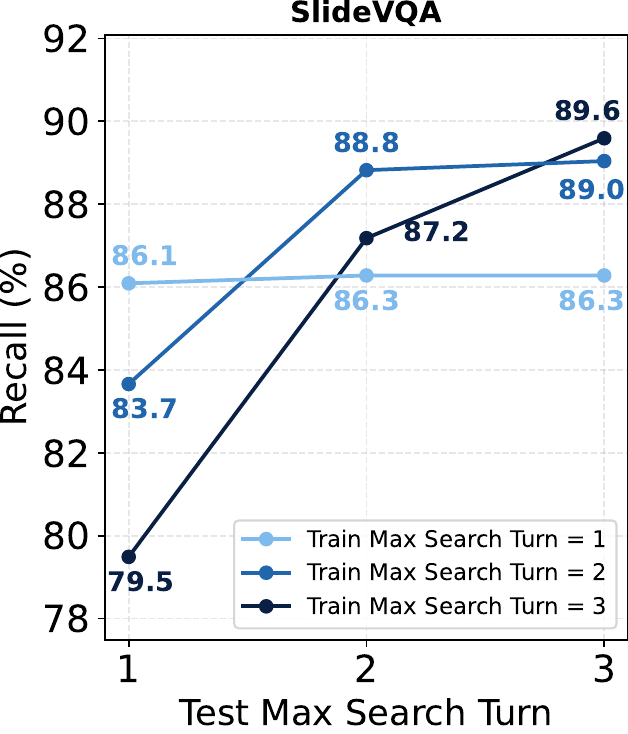}
        \end{minipage}
        \hfill
        \begin{minipage}[t]{0.49\textwidth}
            \centering
            \includegraphics[width=\textwidth]{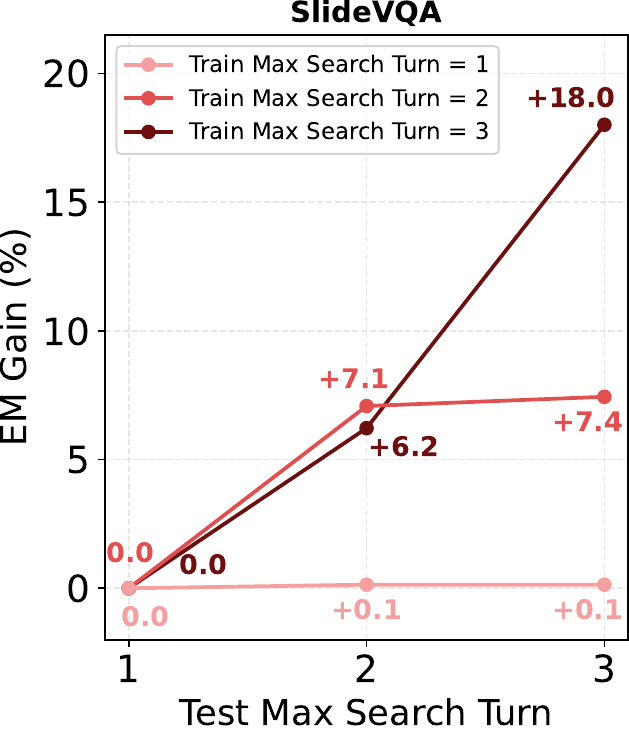}
        \end{minipage}
    \end{minipage}
\vspace{-1mm}
\caption{Search turn discussion on multi-page scenarios, \emph{i.e.}, MMLongBench-Doc (MP) and SlideVQA at top. 
(1) On both of the datasets, Search performance (Recall) scales consistently with both training (different curve colors) and test-time (from left to right) search budgets. 
(2) For each training-time max search turn (different red colors),  we compute the QA improvement (EM gain) toward the smallest testing-time max search turn. The slopes reflect how effectively the agent converts additional test-time search turns into QA accuracy. Richer training search complexity endows the agent with better capabilities that scale with the test-time search budget.}
\vspace{-5mm}
\label{fig:search_turn_analysis}
\end{figure*}

\textbf{Search Turn Discussion.}
Figure~\ref{fig:search_turn_analysis} examines how the training search turn budget $T$ and the search opportunities at inference time affect the behaviour of the agent. We train variants with $T \in \{1, 2, 3\}$ on SlideVQA and $T \in \{2, 3, 4\}$ on MMLongBench-Doc, and evaluate each under varying test-time search budgets. (1)\textit{Retrieval scaling.} The recall curves show that both higher training turn budgets and more test-time search turns consistently improve the retrieval recall on both benchmarks.
(2) \textit{QA scaling.} The EM gain curves reveal a more nuanced pattern. Each curve plots the QA improvement relative to the minimum test-time search budget, so all curves start at zero and their slopes reflect how effectively the agent converts additional search turns into answer accuracy. On SlideVQA, the $T{=}1$ agent gains only $+0.1$ EM when given two additional test-time search turns. Since it does not practice the multi-round retrieval during training, it fails to leverage the extra search opportunities. The $T{=}2$ agent gains $+7.4$ EM, and the $T{=}3$ agent gains $+18.0$ EM, converting each additional test-time turn into substantial QA improvement. The same pattern holds on MMLongBench-Doc. Training with $T{=}4$ yields a slightly lower gain ($+3.9$) despite achieving the highest recall ($63.9\%$), consistent with a retrieval-reasoning trade-off where excessively aggressive retrieval can introduce contextual noise that mitigate the benefit of higher evidence coverage.

Overall, these results suggest that increasing training search complexity not only improves the absolute QA accuracy, but also alters the agents' behavior and search strategy. Agents trained with richer search budgets may acquire generalizable search-then-reason capabilities that scale with the test-time search budget. Such  a finding reversely validates the effectiveness of the training environment constructed by the DocArena pipeline: the multi-page, cross-document reasoning samples provide a sufficiently complex search space for the agent to develop transferable search strategies.

\section{Conclusion}\label{sec:conclusion}

This work presented DocArena, a fully automated pipeline that transformed raw multimodal document collections into controllable training environments for search agents without human annotation. The pipeline leveraged cross-page information distribution profiling to guarantee evidence exclusivity and cascaded quality assurance to eliminate noise. The resulting DocArena-79K dataset covered 8,336 documents across 16 domains and 49 languages. Together with a Doc-Search agent infrastructure that decoupled visual perception from the policy model, experiments on six multimodal document scenarios and seven text-based QA benchmarks demonstrated the best overall retrieval and QA performance. Analysis of search behaviors further confirmed the controllability of the constructed environment. We hope this work inspires future study toward automated, customizable training environment construction for search agents.



\appendix
\section{Overview}\label{sec:supply_overview}

The appendix is organized as follows:

\medskip
\noindent\textbf{Section~\ref{sec:curation_pipeline}: Data Curation Pipeline}
\begin{itemize}[nosep,leftmargin=2em]
    \item \ref{sec:exclusivity}~More Illustrations on Irreplaceable Evidence
    \item \ref{sec:pipeline_stats}~Pipeline Yield and Rejection Statistics
    \item \ref{sec:cost_analysis}~Computational Cost Analysis
    \item \ref{sec:prompts}~MLLM Prompt Templates
    \item \ref{sec:stage1}~Stage~I: Document Structuring and Indexing
    \item \ref{sec:stage2}~Stage~II: Cross-Page Distribution Profiling
    \item \ref{sec:stage3}~Stage~III: Reasoning Chain Construction
    \item \ref{sec:stage4}~Stage~IV: Cascaded Quality Assurance
    \item \ref{sec:source_collection}~Source Collection and Dataset Statistics
\end{itemize}

\noindent\textbf{Section~\ref{sec:agent_infra}: Doc-Search Agent Infrastructure and Training}
\begin{itemize}[nosep,leftmargin=2em]
    \item \ref{sec:reward_design}~More Illustrations on Reward Design and Credit Assignment
    \item \ref{sec:hyperparams}~Training Hyperparameters
    \item \ref{sec:ocr_tool}~More Illustrations on the Online OCR Tool
    \item \ref{sec:inference_prompt}~More Illustrations on the Inference Prompt Template
\end{itemize}

\noindent\textbf{Section~\ref{sec:eval_setup}: Evaluation Benchmarks and Implementation Details}
\begin{itemize}[nosep,leftmargin=2em]
    \item \ref{sec:bench_mmlongbench}~More Illustrations on MMLongBench-Doc
    \item \ref{sec:bench_visrbench}~More Illustrations on VisRBench
    \item \ref{sec:bench_slidevqa}~More Illustrations on SlideVQA
    \item \ref{sec:baseline_fairness}~More Illustrations on the Comparison
    \item \ref{sec:cross_doc}~Intra/Inter-Document Illustrations
\end{itemize}

\noindent\textbf{Section~\ref{sec:agent_behavior}: Agent Behavior Analysis and Ablation Studies}
\begin{itemize}[nosep,leftmargin=2em]
    \item \ref{sec:data_source_ablation}~Training Data Source Ablation
    \item \ref{sec:data_scaling}~Data Scaling Curves
    \item \ref{sec:base_model}~Base Model Comparison
    \item \ref{sec:nrdup_analysis}~NrDup Metric
\end{itemize}
\newpage

\section{Data Curation Pipeline}\label{sec:curation_pipeline}

\subsection{More Illustrations on Irreplaceable Evidence}\label{sec:exclusivity}

Stage~II selects evidence pages based on the distribution profile and requires that the reasoning chain depends on at least one $w(c){=}1$ fact. Stage~IV then tests the generated QA pair: the MLLM correctness verification (Layer~2) checks whether the answer can be reproduced exactly from the evidence pages, and the leave-one-page-out test (Layer~3) checks whether removing any single evidence page makes the question unanswerable. The two stages work together. 

\subsection{Pipeline Yield and Rejection Statistics}\label{sec:pipeline_stats}

\begin{figure}[t]
\centering
\includegraphics[width=\columnwidth]{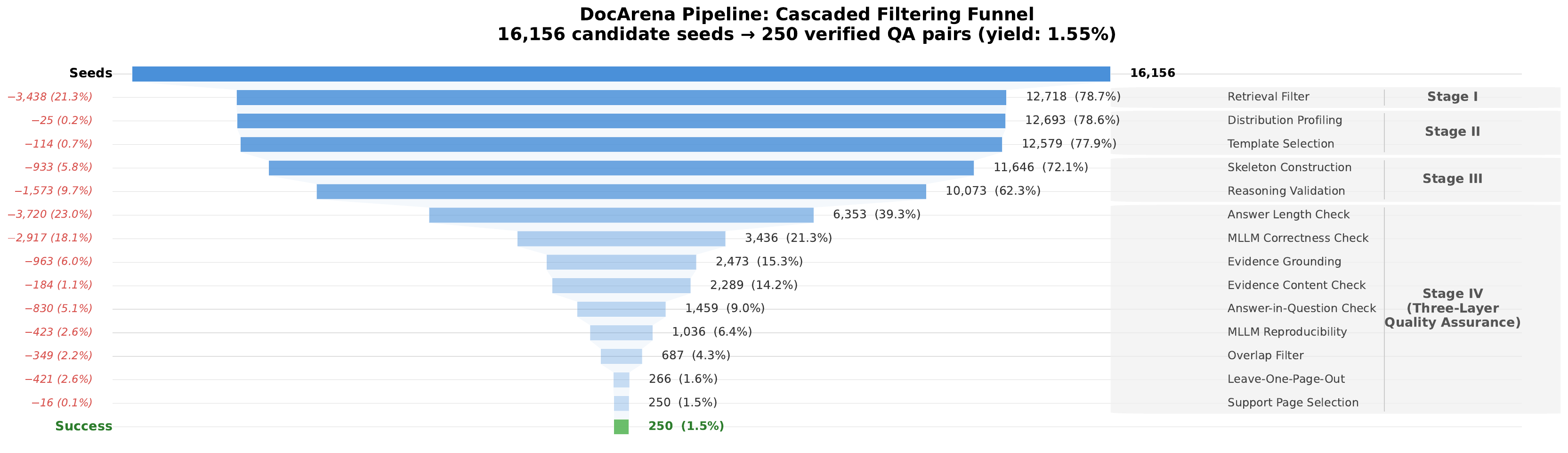}
\vspace{-2mm}
\caption{Cascaded filtering funnel of the DocArena curation pipeline. From $16{,}156$ candidate seeds, each gate progressively filters low-quality samples, yielding $250$ valid QA pairs ($1.55\%$ yield rate). Red annotations indicate the number and percentage of samples rejected at each gate.}
\vspace{-2mm}
\label{fig:pipeline_funnel}
\end{figure}

To characterize the pipeline's filtering behavior, we run the full curation pipeline on a subset and track per-stage outcomes. We generate $250$ verified QA pairs from in total of $16{,}156$ candidate seed pages for analysis (yield rate $1.55\%$, $95\%$ CI: $[1.36\%, 1.74\%]$). Figure~\ref{fig:pipeline_funnel} visualizes the cascaded filtering funnel. The x-axis lists the ten sequential gates grouped by pipeline stage (Stages~II--IV), and the bar height indicates the number of samples surviving after each gate. Red annotations show the count and percentage of samples rejected at each gate. The three largest rejection sources are the answer length filter ($23.0\%$), the retrieval pre-filter ($21.3\%$, documents with fewer than $3$ same-document pages in the retrieval neighborhood), and the MLLM correctness verification ($18.1\%$). The rejection load is distributed across all four stages rather than concentrated in a single gate, indicating that each stage addresses a distinct quality dimension.

\begin{figure}[t]
\centering
\includegraphics[width=0.49\columnwidth]{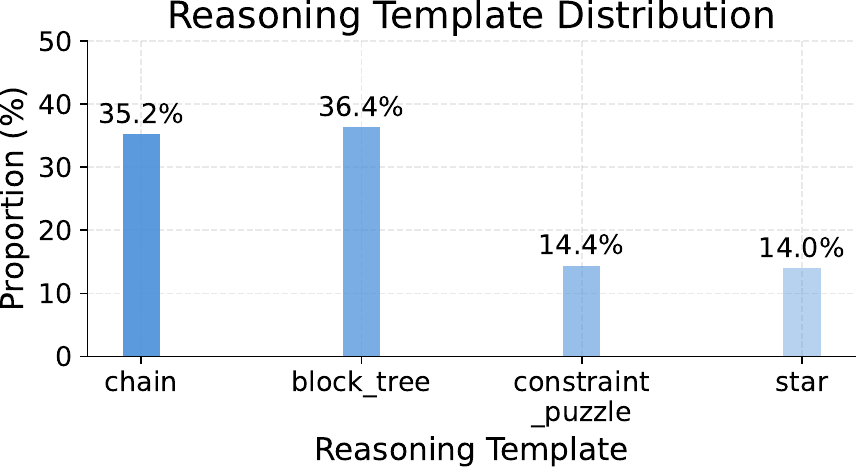}%
\hfill
\includegraphics[width=0.49\columnwidth]{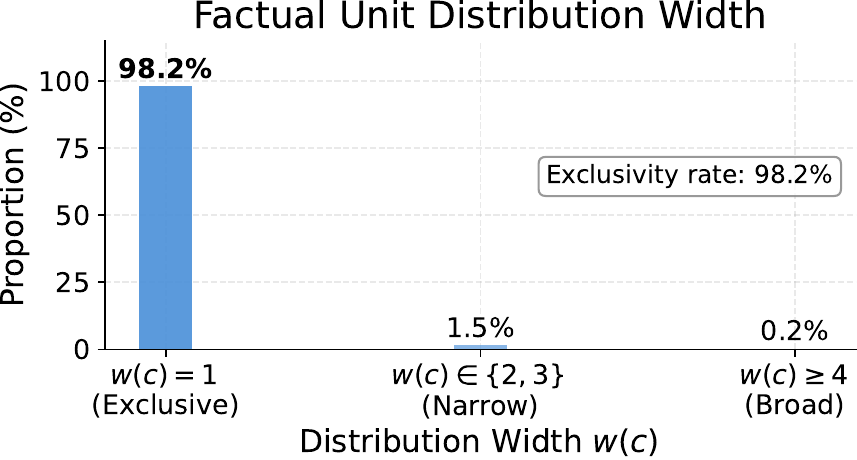}%
\vspace{-2mm}
\caption{\textbf{Left}: Reasoning template distribution among valid QA pairs. \textbf{Right}: Distribution width $w(c)$ of factual units. $98.2\%$ of units are exclusive to a single page ($w{=}1$).}
\vspace{-2mm}
\label{fig:template_width}
\end{figure}

Figure~\ref{fig:template_width} (left) shows the reasoning template distribution among the verified QA pairs. The x-axis lists the four templates and the y-axis shows the proportion. \texttt{chain} ($35.2\%$) and \texttt{block\_tree} ($36.4\%$) together account for over $70\%$ of samples, while \texttt{constraint\_puzzle} ($14.4\%$) and \texttt{star} ($14.0\%$) each contribute a smaller share. No single template exceeds $40\%$, indicating that the proposed design prevents reasoning type collapse. Figure~\ref{fig:template_width} (right) shows the distribution width $w(c)$ of all $22{,}354$ factual units extracted from the verified samples. $98.2\%$ of factual units have $w(c){=}1$ (exclusive to a single page), $1.5\%$ have $w(c) \in \{2,3\}$ (narrow), and $0.2\%$ have $w(c) \geq 4$ (broad). The fact that the majority of factual units are page-exclusive indicates that the distribution profiling stage (Stage~II) identifies \textbf{page-unique} information, providing a reliable foundation for the evidence exclusivity condition described in the main paper.

The statistics reported in this section are based on Qwen2.5-VL-32B-Instruct~\cite{bai2025qwen2} (Apache-2.0 license), which serves as the MLLM backbone throughout the curation pipeline. Essentially, The DocArena pipeline is model-agnostic and readily supports different VLMs as the backbone. However, due to the licensing constraints---GPT-4o is a proprietary closed-source model accessible only through OpenAI's commercial API, and Qwen2.5-VL-72B-Instruct is released under the Qwen Research License Agreement (QRLA), which imposes commercial use restrictions---the publicly released DocArena-79K dataset and all reported statistics in the paper are based exclusively on Qwen2.5-VL-32B-Instruct to ensure full reproducibility and redistribution rights. We note that empirically, the proposed pipeline readily supports different VLMs, and the rejection rate distribution may vary with the backbone model's capability.

\subsection{Computational Cost Analysis}\label{sec:cost_analysis}

\begin{figure}[t]
\centering
\includegraphics[width=0.49\columnwidth]{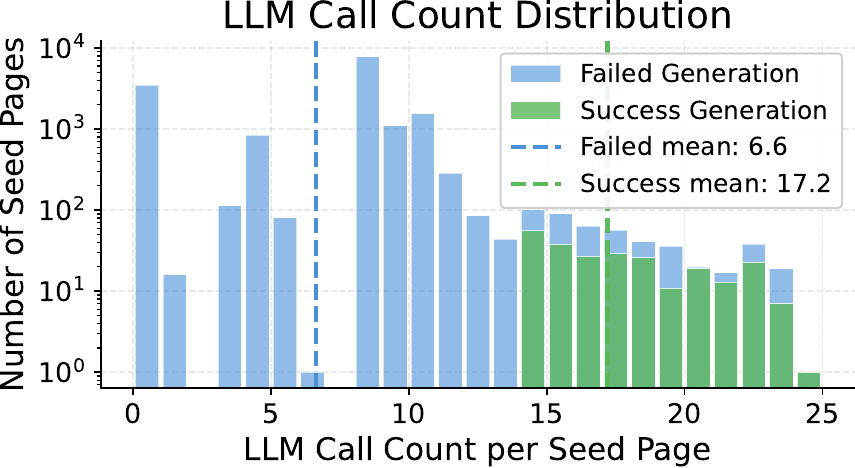}%
\hfill
\includegraphics[width=0.49\columnwidth]{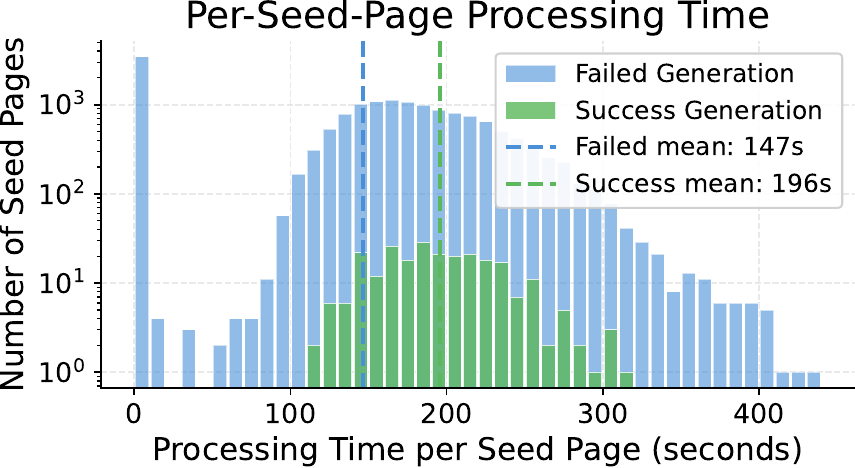}%
\vspace{-2mm}
\caption{\textbf{Left}: MLLM call count per seed page for successful (green) vs.\ failed (blue) generations. Successful seed pages require more calls ($17.2$ mean) as they pass through all cascaded gates. \textbf{Right}: Per-seed-page processing time. The spike near $0$s reflects seed pages rejected by the retrieval pre-filter without any MLLM call. Dashed lines indicate means.}
\vspace{-2mm}
\label{fig:cost_analysis}
\end{figure}

Figure~\ref{fig:cost_analysis} presents the computational cost analysis of the curation pipeline. The left panel shows the distribution of MLLM call counts per seed page (x-axis) against the number of seed pages (y-axis, log scale), split into successful generation (green) and failed generation (blue). Successful seed pages require an average of $17.2$ MLLM calls, while failed seed pages average $6.6$ calls. This gap arises because seed pages that pass all cascaded gates traverse more verification steps, whereas low-quality candidates are rejected early by deterministic filters in Layer~1 before reaching the MLLM-based checks in Layers~2 and~3. The right panel shows the per-seed-page processing time with a similar split. Successful seed pages average $196$s compared to $147$s for failed ones. A visible spike near $0$s corresponds to seed pages rejected by the retrieval pre-filter without any MLLM call. notably, for the failed distribution, its tail extends beyond that of successful cases as some failed seed pages pass the inexpensive deterministic filters in Layer~1 and enter the MLLM-based verification in Layers~2 and~3, accumulating processing time comparable to successful cases before eventually being rejected. Together, the two distributions indicate that the three-layer ordering by computational cost (Section~\ref{sec:stage4}) reduces unnecessary MLLM usage on low-quality candidates while concentrating computation on samples that have a higher chance of passing all gates.

\begin{figure}[t]
\centering
\includegraphics[width=0.49\columnwidth]{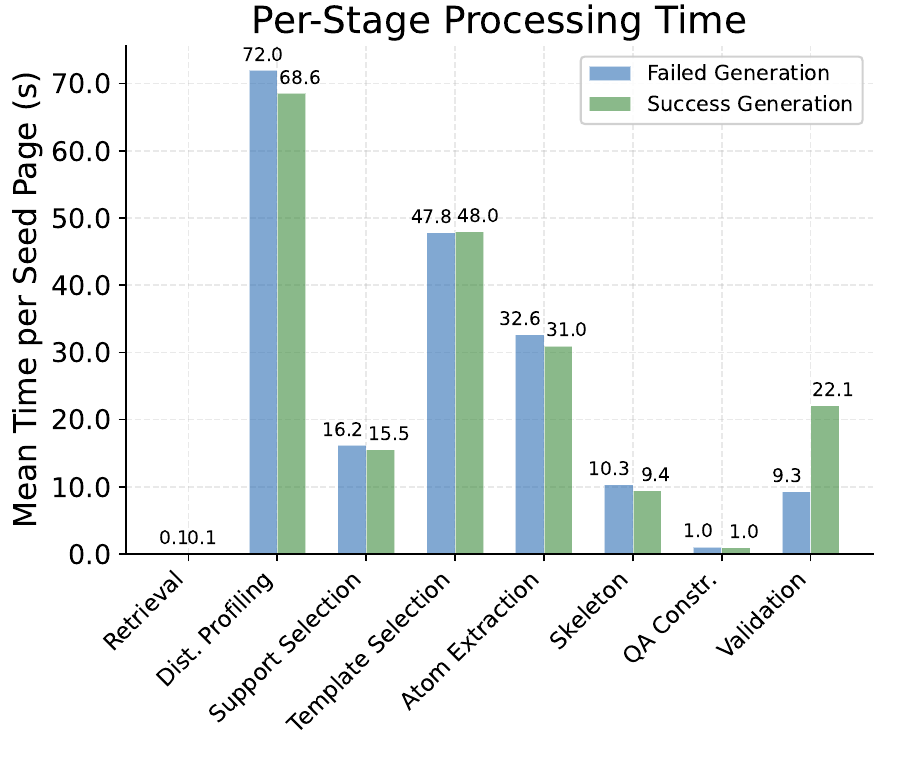}%
\hfill
\includegraphics[width=0.49\columnwidth]{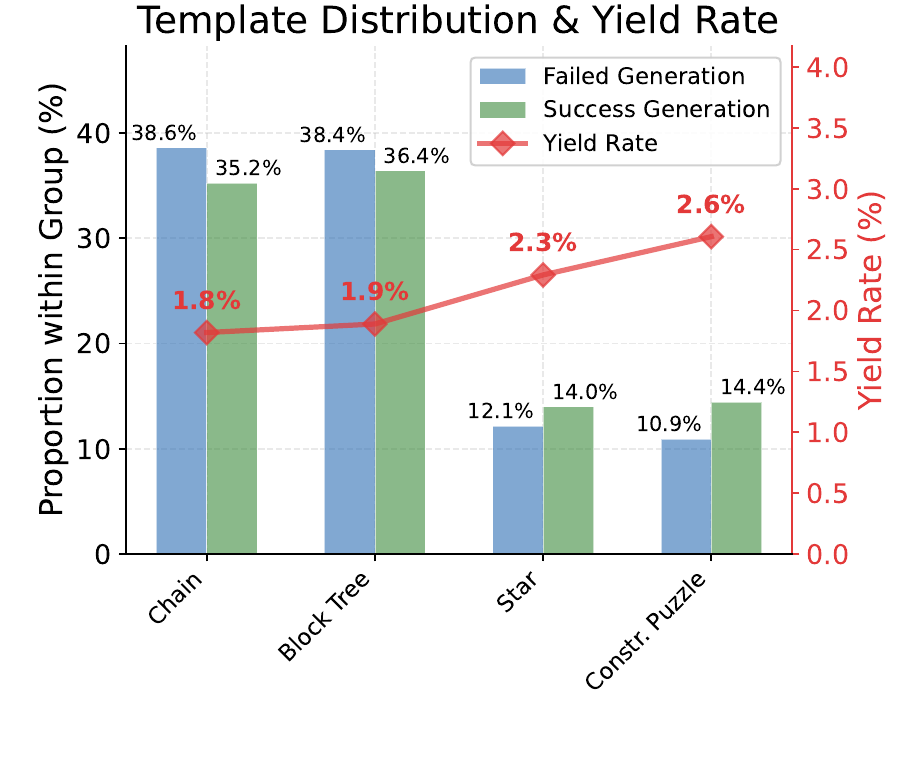}%
\vspace{-2mm}
\caption{\textbf{Left}: Per-seed-page processing time by pipeline stage for successful (green) and failed (blue) generations. Retrieval takes $<\!0.1$s. \textbf{Right}: Proportion of each reasoning template within each outcome group (blue: failed; green: successful), with yield rate per template (red line, right axis). Both success/fail distributions and yield rates ($1.8$--$2.6\%$) are balanced across templates.}
\vspace{-2.5mm}
\label{fig:stage_timing_yield}
\end{figure}

Figure~\ref{fig:stage_timing_yield} breaks down the computational cost by pipeline stage and reasoning template.
The left panel reports the mean processing time of each stage for successful (green) and failed (blue) generations.
Retrieval (Stage~I) takes less than $0.1$s per seed page because it relies on a pre-built FAISS index lookup without any MLLM call.
The four MLLM-intensive stages account for most of the cost: Distribution Profiling (${\sim}70$s), Template Selection (${\sim}48$s), Atom Extraction (${\sim}32$s), and Support Selection (${\sim}16$s).
Across these stages, successful and failed cases show similar mean times, suggesting that the per-stage cost depends on the input complexity rather than on whether the sample eventually passes.
Validation is different: successful cases ($22.1$s) take $2.4{\times}$ longer than failed cases ($9.3$s).
This is expected, as a successful sample must pass all nine cascaded gates in Layers~2 and~3, while a failed sample exits at the first gate it does not pass, and the remaining gates are skipped.

The right panel shows the proportion of each reasoning template within each outcome group: blue bars represent the share of each template among all failed generations, and green bars represent the share among all successful generations. The red line (right axis) shows the yield rate per template.
Both the failed and successful distributions are balanced across templates.
The yield rate is also consistent, ranging from $1.8\%$ to $2.6\%$ across the four templates.
This indicates that the cascaded quality filters do not favor or penalize any particular reasoning structure, and the template diversity in the final dataset reflects the input distribution rather than a filtering bias.

\subsection{MLLM Prompt Templates}\label{sec:prompts}

The curation pipeline (Section~3.2 of the main paper) uses MLLM calls at multiple stages. All calls use Qwen2.5-VL-32B-Instruct~\cite{bai2025qwen2}. We list four representative prompt templates below, organized by pipeline stage. Variables enclosed in braces (e.g., \texttt{\{text\}}) are replaced with the actual content at runtime.

\subsubsection{Stage~II: Factual Unit Extraction.}

For each page in the candidate set $R$, the following prompt extracts structured factual units for cross-page matching.

\begin{tcolorbox}[colback=blue!5, colframe=blue!40!black, title=Factual Unit Extraction Prompt, breakable]
\footnotesize
Extract atomic factual units from this document page.\\[4pt]
PAGE: \{page\_id\}\\
CONTENT: \{text\}\\[4pt]
For each factual unit, provide:\\
1. A concise description (one sentence)\\
2. The type: one of [entity\_name, date, numeric\_value, specification, procedure, requirement, condition, definition, classification, reference, relationship, other]\\
3. A short semantic grouping key for matching equivalent facts across pages (e.g. ``product\_temperature\_range'', ``certification\_standard''). The key must be general enough to match the SAME concept across different pages.\\
4. The specific text/value from this page (brief quote)\\[4pt]
CRITICAL:\\
- Focus on SUBSTANTIVE content, not headers/footers/boilerplate\\
- Extract specific values, named entities, technical specifications, requirements\\
- Grouping keys must be concept-level, not document-specific (e.g. ``operating\_temperature\_range'' not ``model\_x100\_temp\_range'')\\[4pt]
Respond in JSON: \{``factual\_units'': [\{``description'': ..., ``type'': ..., ``grouping\_key'': ..., ``value'': ...\}]\}
\end{tcolorbox}

\subsubsection{Stage~III: Reasoning Skeleton Construction.}

This prompt builds the reasoning program using the selected template, operators, and anchor constraints.

\begin{tcolorbox}[colback=blue!5, colframe=blue!40!black, title=Reasoning Skeleton Construction Prompt, breakable]
\footnotesize
Build a program using the `\{selected\_template\}' template that combines atoms from \{num\_pages\} pages.\\[4pt]
Template: \{selected\_template\}\\
Pages: \{support\_page\_ids\}\\
Extracted atoms: \{atoms\_summary\}\\[4pt]
ANCHOR POINTS (must be used in the reasoning chain): \{anchor\_summary\}\\[4pt]
RELEVANT INFORMATION DISTRIBUTION: \{relevant\_dist\_json\}\\[4pt]
CRITICAL CONSTRAINTS:\\
1. Each operation must use inputs from DIFFERENT pages\\
2. The final answer must require information from ALL pages\\
3. No single page should contain all necessary information\\[4pt]
NUMERICAL CALCULATION REQUIREMENTS:\\
- When building arithmetic operations, explicitly state the exact numbers from the pages\\
- Show step-by-step calculation with actual values (e.g., ``25.5 + 30.2 = 55.7'')\\
- Use only the specific numerical values found in the provided pages\\
- Do not estimate or approximate -- use exact figures from the source material\\[4pt]
Build a program that:\\
- Uses appropriate operators (JOIN, FILTER, AGG, LOOKUP, ARITH)\\
- Ensures cross-page dependencies\\
- Follows the \{selected\_template\} structural pattern\\
- Produces a DERIVED answer that can only be determined by combining information from ALL pages\\
- The expected\_answer must be concise: a specific value, name, identifier, short phrase\\
- The answer should be the result of cross-page information synthesis, not a direct quote from any single page\\[4pt]
Respond with: \{``program\_steps'': [...], ``expected\_answer'': ``...'', ``cross\_page\_verification'': ``...''\}
\end{tcolorbox}

\subsubsection{Stage~III: Question Verbalization.}

This prompt converts the reasoning program into a natural language question with anchor stratification (Section~\ref{sec:stage3}).

\begin{tcolorbox}[colback=blue!5, colframe=blue!40!black, title=Question Verbalization Prompt, breakable]
\footnotesize
Convert this program into a natural language question that would lead to the answer: ``\{answer\}''\\[4pt]
Program steps: \{program\_steps\}\\[4pt]
PRIMARY ANCHOR (mention with specific identifiers): \{primary\_desc\}\\[4pt]
SECONDARY ANCHORS (reference implicitly): \{secondary\_descs\}\\[4pt]
REQUIREMENTS:\\
1. Do NOT mention specific page numbers or locations\\
2. The question should naturally require the information processing described in the program\\
3. Make it sound like a realistic question someone would ask\\
4. Ensure the question implies the need for multiple pieces of information\\[4pt]
Generate a clear, natural question.
\end{tcolorbox}

\subsubsection{Stage~IV Layer~3: Leave-One-Page-Out Test.}

This prompt tests whether the question remains answerable when each evidence page is excluded.

\begin{tcolorbox}[colback=blue!5, colframe=blue!40!black, title=Leave-One-Page-Out Test Prompt, breakable]
\footnotesize
Based ONLY on the following pages, answer this question.\\
If the pages contain enough information to answer fully, provide a brief answer.\\
If the pages do NOT contain enough information, respond with exactly: UNANSWERABLE\\[4pt]
Pages: \{remaining\_pages\_text\}\\[4pt]
Question: \{question\}\\[4pt]
Answer (or UNANSWERABLE):
\end{tcolorbox}

\subsection{Stage~I: Document Structuring and Indexing}\label{sec:stage1}

Each page is rendered at 144~DPI, matching the page-level input granularity of the ColPali retriever used during RL training.

The dense retrieval index in Stage~I uses E5-base-v2~\cite{wang2022text} as the text encoder. This index serves the pipeline's internal cross-page retrieval needs in Stage~II and is separate from the ColPali retriever used by the Doc-Search agent at training and inference time. During data curation, the pipeline operates on structured semantic text extracted by the MLLM, so a text encoder can capture fine-grained semantic similarity between pages. During agent interaction, the retriever operates on raw page images and the agent's generated queries, so ColPali's vision-language alignment is more suitable. This separation also avoids coupling the pipeline's retrieval quality with the agent's retrieval quality: the pipeline index can be built once, while the agent retriever can be changed independently without re-running the curation pipeline.

\subsection{Stage~II: Cross-Page Distribution Profiling}\label{sec:stage2}

Given a seed page, the pipeline retrieves from the global FAISS index and then filters to retain pages from the same document as the seed page (Section~\ref{sec:cross_doc}). The top $K$ pages filtering form the candidate set $R$. A subset of up to $15$ pages from $R$ is selected for distribution analysis to balance coverage and MLLM cost.

Documents are skipped if fewer than $3$ same-document pages are retrieved or if fewer than $2$ pages yield factual units. After aggregation, documents with no exclusive ($w{=}1$) or narrow ($w \in \{2,3\}$) concepts are also skipped, as the distribution profile does not contain sufficient structure to support evidence-exclusive QA construction.

\subsection{Stage~III: Reasoning Chain Construction}\label{sec:stage3}

The reasoning skeleton is constructed using five operators:
\begin{itemize}
  \item \texttt{LOOKUP}: extract a value from a specific page.
  \item \texttt{JOIN}: associate information across pages.
  \item \texttt{FILTER}: apply a condition to select relevant entries.
  \item \texttt{AGG}: aggregate values (sum, average, count).
  \item \texttt{ARITH}: perform arithmetic ($+$, $-$, $\times$, $\div$, $\%$).
\end{itemize}
Each operator takes inputs from different pages, and the skeleton enforces that the final answer requires information from all selected evidence pages. For templates involving numerical calculations, the skeleton includes step-by-step computation with exact values from the pages.

During QA verbalization, the pipeline stratifies anchor points into a primary anchor (mentioned with specific identifiers in the question) and secondary anchors (referenced implicitly). The primary anchor is selected as the one with the lowest distribution width and the highest discriminative score, which is computed based on the presence of named entities, clause or section numbers, and numerical values in the anchor description. This stratification ensures that the question contains a concrete search clue from the irreplaceable evidence without revealing the answer.

The minimum number of evidence pages per sample is configurable between $2$ and $4$ (default $2$). Answer length is constrained to at most $15$ words or $80$ characters.

\subsection{Stage~IV: Cascaded Quality Assurance}\label{sec:stage4}

\subsubsection{More Discussions on MLLM Error Propagation.}
The curation pipeline uses MLLM calls in Stages~II and~III to extract factual units and construct reasoning chains. Errors can be introduced during these stages, and  may be propagated into the generated QA pairs. The three-layer cascaded quality assurance in Stage~IV is designed to catch  errors. Layer~1 applies deterministic filters that check whether the answer is grounded in the evidence text; QA pairs where the answer contains facts not present in the evidence pages are rejected. Layer~2 uses a separate MLLM call to re-answer the question given only the evidence pages; if the reproduced answer is inconsistent with the curated answer, the sample is rejected, which catches cases where extraction errors led to an unanswerable or incorrectly answered question. Layer~3 tests whether each evidence page is necessary; extraction errors that cause the pipeline to include a redundant page are caught when that page can be removed without affecting answerability. The three layers are ordered by increasing computational cost, so that the deterministic filters in Layer~1 remove low-quality samples early and reduce the number of MLLM calls needed in Layers~2 and~3.

\subsection{Source Collection and Dataset Statistics}\label{sec:source_collection}

CCpdf~\cite{turski2023ccpdf} is a large-scale, multilingual corpus of visually rich PDF documents sourced from Common Crawl. The corpus contains 1.1 million documents with 14.5 million pages, with an average of 12.9 pages per document. The documents are sourced from diverse websites across the Internet, covering business, legal, scientific, and technical domains, with the majority created after 2010. Compared to prior publicly available document corpora such as IIT-CDIP (6.5M documents from a single domain in the 1990s) and OCR-IDL (4.6M single-domain documents), CCpdf provides broader domain and language coverage with longer documents on average. The corpus is released as an open-license index with downloading scripts, and the pipeline can be applied to additional Common Crawl dumps for further scaling. We select CCpdf as the source collection for DocArena because its domain diversity, language coverage, and visual richness (tables, figures, charts, structured layouts) provide the variety needed to construct a training environment that generalizes across document types.

\begin{figure}[t]
\centering
\includegraphics[width=0.45\columnwidth]{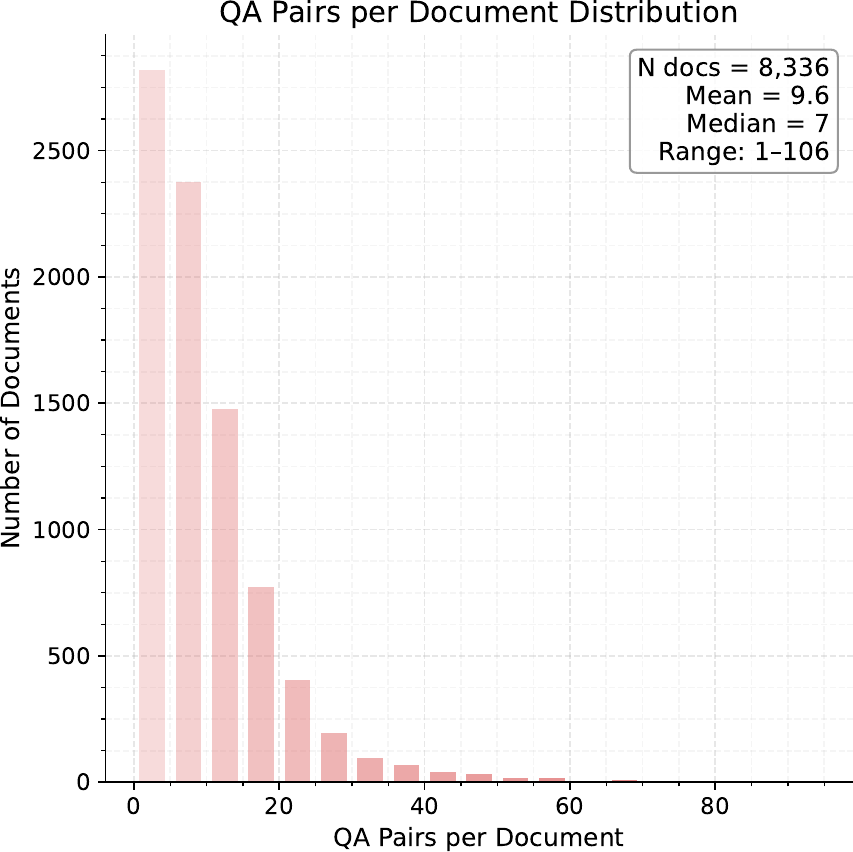}
\vspace{-2mm}
\caption{Distribution of QA pairs per document in DocArena-79K (mean $9.6$, median $7$).}
\vspace{-2mm}
\label{fig:extra_stats}
\end{figure}

Figure~\ref{fig:extra_stats} presents the distribution of QA pairs per document in DocArena-79K. Across $8{,}336$ source documents, the pipeline produces an average of $9.6$ QA pairs per document (median $7$), with a long tail extending up to $106$ pairs for documents with rich cross-page information structure.

\section{Doc-Search Agent Infrastructure and Training}\label{sec:agent_infra}

\subsection{More Illustrations on Reward Design and Credit Assignment}\label{sec:reward_design}

As stated in Eq.~3 of the main paper, the total reward $R(\tau)$ combines a QA reward $r^{\text{QA}}$ and a progress reward $r^{\text{PR}}_t$. 

\paragraph{Progress reward computation.} At each search turn $t$, the agent retrieves a set of pages $\mathcal{R}_t$. The progress reward is the recall of the retrieved pages against the ground-truth evidence pages $\mathcal{E}$:
\begin{equation}
r^{\text{PR}}_t = \frac{|\mathcal{R}_t \cap \mathcal{E}|}{|\mathcal{E}|}.
\end{equation}

\paragraph{Credit assignment.} The progress reward is assigned to the search action token for fine-grained credit assignment. Specifically, the discounted progress reward (with $\gamma{=}0.9$) is placed at the token position corresponding to the end of the search action tag (\texttt{</search>}) at turn $t$, and the QA reward $r^{\text{QA}}$ is placed at the last token of the response sequence. This token-level placement allows the policy gradient to attribute retrieval behavior and answer generation to their respective action tokens within the same trajectory.

\subsection{Training Hyperparameters}\label{sec:hyperparams}

Table~\ref{tab:hyperparams} provides the complete set of training hyperparameters used for the Doc-Search agent. The implementation is built on veRL~\cite{sheng2024hybridflow} with vLLM for rollout generation.

\begin{table}[h]
\centering
\caption{Training hyperparameters for the Doc-Search agent.}
\label{tab:hyperparams}
\resizebox{0.35\columnwidth}{!}{
\begin{tabular}{l|l}
\hline
\textbf{Parameter} & \textbf{Value} \\
\hline
Base model & Qwen2.5-7B-Instruct \\
Optimizer & GRPO \\
Batch size & 512 \\
Group size & 5 \\
Learning rate & $1 \times 10^{-6}$ \\
Warmup ratio & 0.285 \\
Total training steps & 1{,}005 \\
Max prompt length & 8{,}192 \\
Max response length & 500 \\
Max start length & 2{,}048 \\
Max observation length & 1{,}000 \\
Max search turns (training) & 2 \\
Retriever top-$K$ & 3 \\
Rollout temperature & 1.0 \\
$\lambda_{\text{QA}}$ & 0.3 \\
$\lambda_{\text{PR}}$ & 0.7 \\
Progress reward weight & 0.5 \\
Discount factor $\gamma$ & 0.9 \\
KL loss coefficient & 0.001 \\
KL loss type & low\_var\_kl \\
Gradient checkpointing & True \\
State masking & True \\
\hline
\end{tabular}
}\vspace{-2mm}
\end{table}

\subsection{More Illustrations on the Online OCR Tool}\label{sec:ocr_tool}

The online OCR tool mentioned in Section~3.4 of the main paper is a modular component of the Doc-Search agent infrastructure that converts retrieved page images into text for the policy model. The specific text extraction method is selected based on the data format of each benchmark: for MMLongBench-Doc, text is extracted from the source PDF using PyMuPDF; for SlideVQA, where documents are slide images without embedded text layers, EasyOCR is used; for VisRBench, pre-processed structured markdown files are used. This per-benchmark adaptation is possible because the Doc-Search agent infrastructure treats the OCR tool as a pluggable module: the policy model receives extracted text regardless of how it was produced, and the same agent checkpoint can be deployed with different extraction methods without retraining. The extracted text is truncated to the max observation length before being appended to the agent's context. Note that for each benchmark, we adopt the same OCR tool for all different baselines for the fair comparison.

\subsection{More Illustrations on the Inference Prompt Template}\label{sec:inference_prompt}

All methods in our evaluation use the same inference prompt template, which follows the format introduced in Search-R1~\cite{jin2025search} as shown below.

\begin{tcolorbox}[colback=blue!5, colframe=blue!40!black, title=Inference Prompt Template, breakable]
\footnotesize
Answer the given question. You must conduct reasoning inside <think> and </think> first every time you get new information. After reasoning, if you find you lack some knowledge, you can call a search engine by <search> query </search> and it will return the top searched results between <information> and </information>. You can search as many times as your want. If you find no further external knowledge needed, you can directly provide the answer inside <answer> and </answer>, without detailed illustrations. For example, <answer> Beijing </answer>.\\[4pt]
Question: \{question\}
\end{tcolorbox}

\section{Evaluation Benchmarks and Implementation Details}\label{sec:eval_setup}

\subsection{More Illustrations on MMLongBench-Doc}\label{sec:bench_mmlongbench}

MMLongBench-Doc~\cite{ma2024mmlongbench} contains $1{,}082$ expert-annotated questions over $135$ long PDF documents with an average of approximately $47.5$ pages per document. The benchmark includes both single-page (SP) questions, where the evidence comes from one page, and multi-page (MP) questions, where the evidence spans multiple pages. Questions labeled as unanswerable in the original benchmark are excluded during data preparation; the evaluation is conducted on the remaining answerable questions only, and all methods are evaluated on the same question set.

\subsection{More Illustrations on VisRBench}\label{sec:bench_visrbench}

VisRBench~\cite{chen2025visr} provides three sub-scenarios defined by the type of visual element that the question targets: FigureQA (questions on figures and charts), TableQA (questions on tables), and TextQA (questions on textual content). All three sub-scenarios are single-page: each question is associated with one evidence page. For each sub-scenario, the original benchmark provides QA pairs in separate JSON files (\texttt{figure\_QA.json}, \texttt{table\_QA.json}, \texttt{text\_QA.json}), each containing per-page question lists with English-translated questions and answers. After excluding samples with empty or unanswerable answers, the evaluated set contains $417$ FigureQA, $4{,}835$ TableQA, and $7{,}268$ TextQA pairs ($12{,}520$ total). We report retrieval and QA metrics separately for each sub-scenario (Tables~3 and~4 of the main paper).

\subsection{More Illustrations on SlideVQA}\label{sec:bench_slidevqa}

SlideVQA~\cite{tanaka2023slidevqa} provides $2{,}215$ multi-page questions over slide presentation decks. Each question requires information from multiple slides within the same deck, and the evidence slide annotations are provided by the original benchmark. Samples labeled as unanswerable are excluded during data loading. 

\subsection{More Illustrations on the Comparison}\label{sec:baseline_fairness}

All compared methods share the same Doc-Search agent infrastructure described in Section~3.4 of the main paper. At inference time, every method uses the same ColPali retriever with top-3 retrieval per query, the same online OCR tool, the same prompt template, and the same generation configurations including max prompt length (8,192 tokens), max response length (1,000 tokens), and max observation length (2,500 tokens). The max observation length controls how many tokens from each retrieval result are kept after OCR conversion. The max search turns is set to 2 for single-page scenarios and 3 for multi-page scenarios for all methods. During training, all methods also share the same set of infrastructure parameters: batch size of 512, max prompt length of 8,192 tokens, max response length of 500 tokens, max start length of 2,048 tokens, max observation length of 1,000 tokens, group size of 5, and max search turns of 2. The max start length controls the token budget for the initial prompt before any search action, and the max observation length during training controls the token budget for each retrieval observation appended to the context. The only variable across methods is the policy model checkpoint.

The fairness of this comparison comes from the infrastructure design. The Doc-Search agent infrastructure converts all multimodal signals (page images, tables, figures, layouts) into text through the ColPali retriever and online OCR tool before passing them to the policy model. The policy model receives and generates text only. This means all compared methods, including those trained on text-based QA data, operate in the same text-in text-out format at both training and inference time. The infrastructure does not require any method to handle visual inputs, and no method has an advantage or disadvantage due to modality differences. The performance differences across methods therefore reflect the differences in their training data and learned search policies, not differences in their ability to process multimodal content.

\subsection{Intra/Inter-Document Illustrations}\label{sec:cross_doc}

The curation pipeline builds a global FAISS index over all pages from the entire document collection during Stage~I (Section~3.2 of the main paper). In Stage~II, given a seed page, the retrieval step returns pages ranked by semantic similarity from this global index, which may come from \textbf{different} documents. The current pipeline  applies a single-document filter that retains only pages belonging to the same document as the seed page (see Fig.~2 of the main paper). Removing this filter is the {only} change needed to enable cross-document QA construction as the remaining pipeline stages (distribution profiling, reasoning chain construction, and cascaded quality assurance) operate on the retrieved page set regardless of document boundaries. Overall,  the proposed pipeline can construct QA pairs where evidence pages span multiple documents when the document collection contains semantically related documents. Notably, the DocArena-79K dataset in this work applies the single-document filter, so all evidence pages $\mathcal{E}$ in each training tuple belong to the same document, and the selected pages span at least $M$ distinct pages within that document (Section~3.2 of the main paper, Fig.~2).

At both training and inference time, the ColPali retriever receives a query and searches across all pages in the document collection without document-level restriction (Section~3.4 of the main paper). The retriever returns the top-$K$ pages ranked by relevance. These pages can come from any document in the collection. By searching across the full collection, the agent faces a challenging retrieval setting that mirrors practical document search scenarios, where the target document is not known in advance.

\section{Agent Behavior Analysis and Ablation Studies}\label{sec:agent_behavior}

\subsection{Training Data Source Ablation}\label{sec:data_source_ablation}

Table~\ref{tab:data_source} compares agents trained with Search-R1 data (NQ + HotpotQA text QA pairs)~\cite{jin2025search} and DocArena-79K under the same Doc-Search agent infrastructure, \emph{e.g.}, Qwen2.5-7B-Instruct, ColPali retriever, training schedule, and identical reward configuration (For the Search-R1 data, the progress reward computes recall against the gold passages provided in the original NQ/HotpotQA annotations).

DocArena-trained agent outperforms the Search-R1-data-trained agent  across both benchmarks. On MMLongBench-Doc MP, DocArena achieves Recall $61.38$ ($+3.83$) and EM $24.18$ ($+4.10$). On SlideVQA MP, DocArena achieves F1 $66.95$ ($+2.76$) and EM $47.96$ ($+3.17$). These gaps isolate the effect of training data quality, thus the performance differences can be mainly attributed to the properties of the training environment---evidence exclusivity, reasoning diversity, and domain coverage.

Notably, the Search-R1-data-trained agent under our infrastructure (Recall $57.55$, EM $20.08$ on MMLongBench-Doc MP) achieves results close to the pretrained Search-R1 model evaluated under the same infrastructure (Recall $55.24$, EM $20.10$ in Table~5 in the main paper). The near-identical EM ($20.08$ vs.\ $20.10$) suggests that the Search-R1 training data reaches its performance ceiling, and the gap relative to DocArena is not due to insufficient training but rather reflects the upper bound of the training data itself.

\begin{table*}[h]
\caption{Training data source ablation on MMLongBench-Doc MP and SlideVQA MP. Both agents use the same Doc-Search agent infrastructure and differ only in training data. Search-R1 data consists of NQ + HotpotQA (${\sim}$169K text QA pairs).}
\vspace{-2mm}
\resizebox{\textwidth}{!}{
\begin{tabular}{c|ccc|ccc|ccc|ccc}
\hline
\multirow{3}{*}{Training Data}
& \multicolumn{6}{c|}{MMLongBench-Doc (MP)}
& \multicolumn{6}{c}{SlideVQA (MP)} \\
\cline{2-13}
& \multicolumn{3}{c|}{Retrieval}
& \multicolumn{3}{c|}{QA Metrics}
& \multicolumn{3}{c|}{Retrieval}
& \multicolumn{3}{c}{QA Metrics} \\
\cline{2-13}
& Recall $\uparrow$ & Prec $\uparrow$ & F1 $\uparrow$
& EM $\uparrow$ & Model $\uparrow$ & PNLS $\uparrow$
& Recall $\uparrow$ & Prec $\uparrow$ & F1 $\uparrow$
& EM $\uparrow$ & Model $\uparrow$ & PNLS $\uparrow$ \\
\hline
Search-R1 Data (Our Infra)
& 57.55 & 34.32 & 39.53
& 20.08 & 19.20 & 28.80
& 85.74 & 56.43 & 64.19
& 44.79 & 42.20 & 50.46 \\
DocArena-79K (Our Infra)
& 61.38\textsubscript{\color{red}+3.83} & 33.76 & 39.68\textsubscript{\color{red}+0.15}
& 24.18\textsubscript{\color{red}+4.10} & 19.43\textsubscript{\color{red}+0.23} & 33.12\textsubscript{\color{red}+4.32}
& 86.28\textsubscript{\color{red}+0.54} & 59.78\textsubscript{\color{red}+3.35} & 66.95\textsubscript{\color{red}+2.76}
& 47.96\textsubscript{\color{red}+3.17} & 42.43\textsubscript{\color{red}+0.23} & 52.18\textsubscript{\color{red}+1.72} \\
\hline
\end{tabular}
}\vspace{-2.5mm}
\label{tab:data_source}
\end{table*}

\subsection{Data Scaling Curve}\label{sec:data_scaling}

We study the scaling behaviour of the DocArena-79K by training separate agents on random subsets of 10\%, 25\%, 50\%, 75\%, and 100\% of the full training set.

\begin{figure}[t]
\centering
\includegraphics[width=0.32\columnwidth]{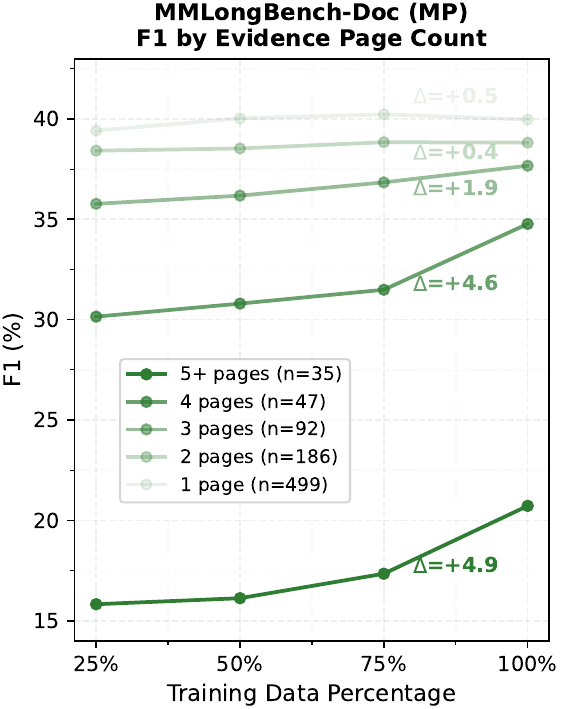}%
\hfill
\includegraphics[width=0.32\columnwidth]{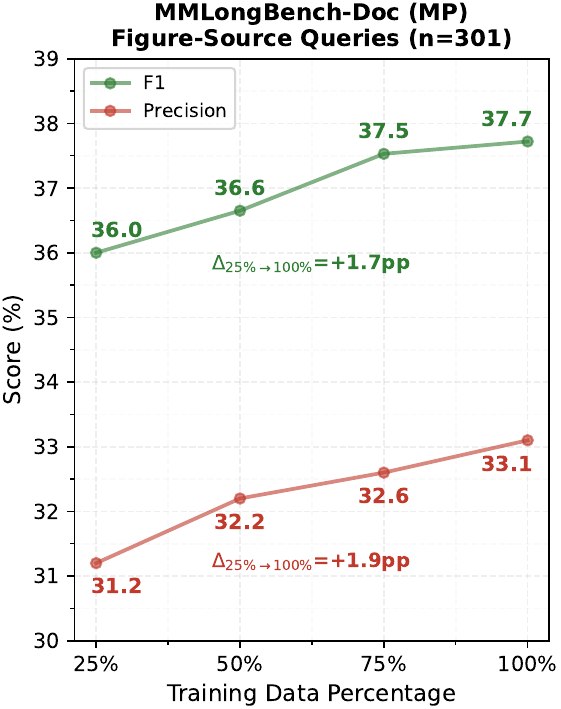}%
\hfill
\includegraphics[width=0.32\columnwidth]{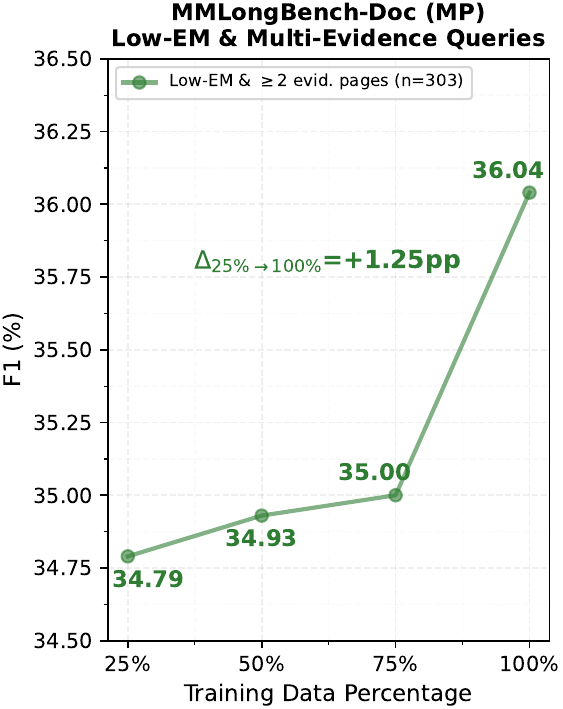}%
\vspace{-2mm}
\caption{Data scaling analysis on MMLongBench-Doc MP. \textbf{Left}: F1 by evidence page count. \textbf{Middle}: F1/Precision on figure-source queries. \textbf{Right}: F1 on low-EM ($\leq$0.5) multi-evidence queries.}
\label{fig:scaling_mmlongbench}
\end{figure}

Fig.~\ref{fig:scaling_mmlongbench} analyzes data scaling on MMLongBench-Doc MP from three representative cases. The left panel plots F1 at each data fraction for queries grouped by evidence page count. All five curves rise from 25\% to 100\%, and the gain grows with retrieval difficulty. Queries requiring 5+ pages improve by $+4.9$ F1, whereas 1-page queries improve by $+0.5$. This gap indicates that the training data contributes more to queries that require retrieving and reasoning over multiple pages. The middle panel reports F1 and Precision specifically on figure-source queries (n=301). Both metrics increase across all four data fractions ($\Delta$F1$=+1.7$, $\Delta$Precision$=+1.9$), showing that the scaling trend holds on queries grounded in \textbf{visual elements}. The right panel selects queries with EM$\leq$0.5 at full data and $\geq$2 evidence pages (n=303). F1 on this subset increases from 34.79 to 36.04 ($+1.25$), showing that more training data improves performance on challenging queries.

\begin{figure}[t]
\centering
\includegraphics[width=0.32\columnwidth]{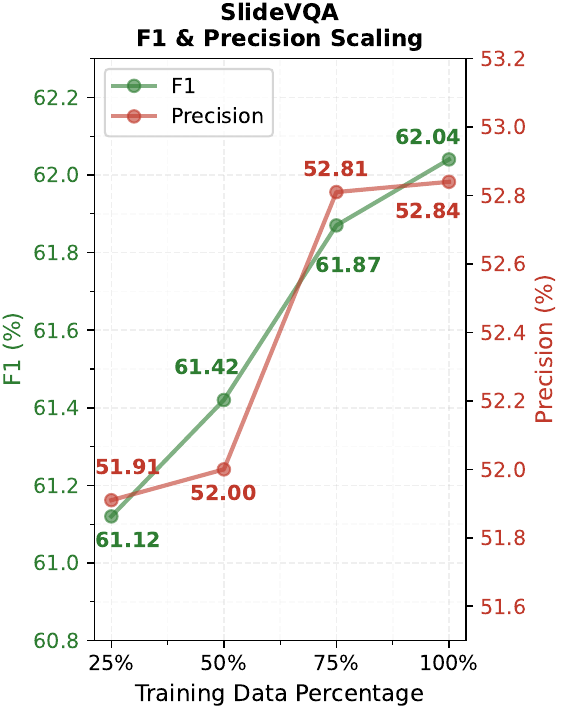}%
\hfill
\includegraphics[width=0.32\columnwidth]{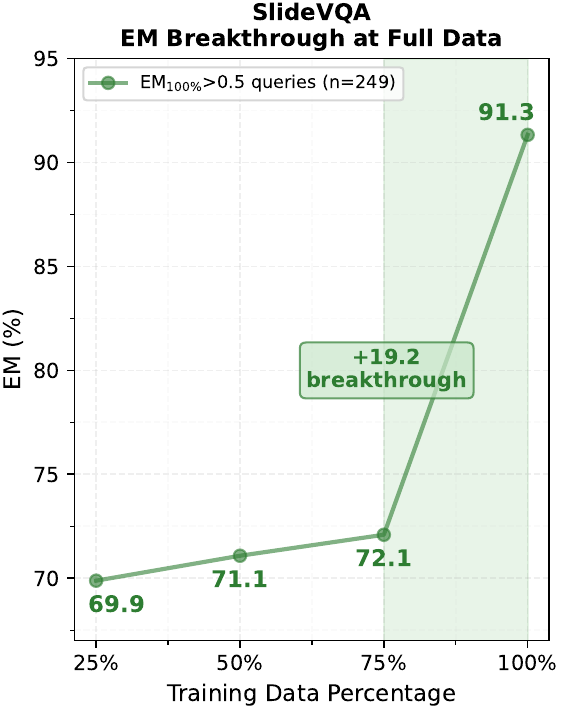}%
\hfill
\includegraphics[width=0.32\columnwidth]{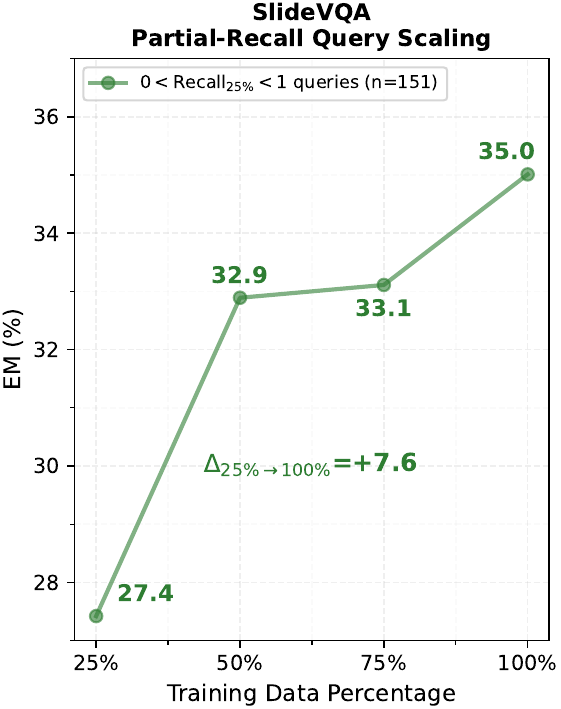}%
\vspace{-2mm}
\caption{Data scaling analysis on SlideVQA MP. \textbf{Left}: F1 and Precision scaling. \textbf{Middle}: EM on queries where the full-data agent achieves EM${>}0.5$ (which are partially solved queries at the boundary of the agent's capability where more training data is more likely to make a difference). \textbf{Right}: EM on queries with partial retrieval recall at the 25\% data level.}
\vspace{-3mm}
\label{fig:scaling_slidevqa}
\end{figure}

Fig.~\ref{fig:scaling_slidevqa} presents scaling results on SlideVQA. The left panel uses dual y-axes to display F1 (left) and Precision (right) at each data fraction. Both metrics rise from 25\% to 100\%, showing a consistent scaling trend across two retrieval metrics. The middle panel examines queries where the full-data agent achieves EM${>}0.5$ (n=249). These are partially solved queries that sit at the boundary of the agent's capability, where additional training data is more likely to push performance from partial to consistent success. EM rises from 69.9 at 25\% to 72.1 at 75\%, and then jumps to 91.3 at 100\%---a $+19.2$ percentage point gain in the last data fraction. This suggests that the agent reaches a threshold where sufficient training diversity enables it to solve a large portion of queries. The right panel tracks queries where the 25\%-data agent retrieves some but not all evidence pages ($0<$Recall$_{25\%}<1$). These queries represent cases where the agent has partial retrieval ability but lacks the search strategy to cover all evidence. EM on this subset improves from 27.4 to 35.0 ($\Delta=+7.6$), indicating that more training data helps the agent learn search strategies that retrieve the remaining evidence pages.

\begin{figure}[h]
\centering
\includegraphics[width=0.4\columnwidth]{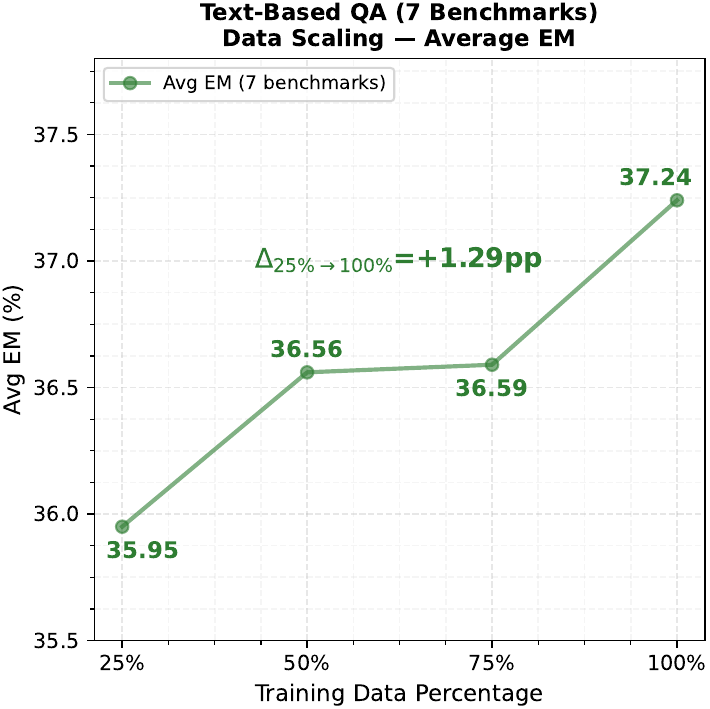}
\caption{Data scaling on text-based QA benchmarks. Average EM across seven benchmarks (Natural Questions, TriviaQA, PopQA, HotpotQA, 2WikiMultiHopQA, MuSiQue, Bamboogle).}
\vspace{-7mm}
\label{fig:scaling_flashrag}
\end{figure}

\begin{table*}[h]
\caption{Data scaling on \textbf{text-based QA benchmarks}. All methods use E5-base-v2 with Wikipedia-18 as the retriever, identical inference parameters, and differ only in the policy model. The upper block lists existing methods for reference; the lower block (\colorbox{gray!48}{shaded}) shows our Doc-Search agent upon different training data fraction of DocArena-79K. The 100\% row corresponds to the result reported in the main paper.}
\vspace{-2mm}
\centering
\resizebox{\textwidth}{!}{
\begin{tabular}{l|ccccccc|c}
\hline
Method
& NQ $\uparrow$ & TriviaQA $\uparrow$ & PopQA $\uparrow$ & HotpotQA $\uparrow$ & 2WikiMultiHopQA $\uparrow$ & MuSiQue $\uparrow$ & Bamboogle $\uparrow$ & Avg $\uparrow$ \\
\hline
Search-R1~\cite{jin2025search}
& 39.20 & 59.22 & 38.91 & 36.16 & 27.58 & 14.94 & 36.29 & 36.04 \\
StepSearch~\cite{wang2025stepsearch}
& 36.84 & 57.51 & 36.41 & 36.31 & 32.05 & \textbf{19.53} & 35.48 & 36.30 \\
ZeroSearch~\cite{sun2025zerosearch}
& 36.29 & 57.09 & 35.79 & 32.11 & 34.90 & 10.88 & 29.03 & 33.73 \\
R-Search~\cite{zhao2025r}
& 34.71 & 58.85 & 36.17 & 32.59 & 29.76 & 12.83 & 35.48 & 34.34 \\
ReSearch~\cite{chen2025learning}
& 36.40 & 58.71 & 38.87 & 34.88 & 27.97 & 17.09 & \textbf{40.32} & 36.32 \\
DeepResearcher~\cite{zheng2025deepresearcher}
& 35.60 & 58.44 & 37.06 & 34.85 & 31.39 & 14.40 & 39.52 & 35.89 \\
\hline
\rowcolor{gray!48} Ours (25\%)  & 36.15 & 59.54 & 39.39 & 35.56 & 33.87 & 13.24 & 33.87 & 35.95 \\
\rowcolor{gray!48} Ours (50\%)  & 38.01 & 61.09 & 40.95 & 34.29 & 34.14 & 12.74 & 34.68 & 36.56 \\
\rowcolor{gray!48} Ours (75\%)  & 36.09 & 59.77 & 40.09 & 37.23 & 36.35 & 13.53 & 33.06 & 36.59 \\
\rowcolor{gray!48} Ours (100\%) & \textbf{36.68} & \textbf{59.63} & \textbf{40.34} & \textbf{37.23} & \textbf{36.82} & 14.11 & 33.87 & \textbf{37.24} \\
\hline
\end{tabular}
}\vspace{-3mm}
\label{tab:scaling_flashrag}
\end{table*}

Fig.~\ref{fig:scaling_flashrag} and Table~\ref{tab:scaling_flashrag} present data scaling results on seven text-based QA benchmarks. Even at 25\% of the training data, our method already surpasses Search-R1, ZeroSearch, R-Search, and DeepResearcher in average EM, suggesting that DocArena-79K provides effective training signal at small scale. As the data fraction increases from 25\% to 100\%, the average EM rises from 35.95 to 37.24. On multi-hop benchmarks such as 2WikiMultiHopQA and HotpotQA, the improvement from 25\% to 100\% reaches $+2.95$ and $+1.67$ respectively. This gap indicates that multi-hop reasoning benefits more from additional training data, as the agent needs to learn search strategies that chain multiple retrieval steps.

\subsection{Base Model Comparison}\label{sec:base_model}

Table~\ref{tab:base_model} compares agents trained from Qwen2.5-7B (base) and Qwen2.5-7B-Instruct on MMLongBench-Doc MP and SlideVQA. Both agents use the same DocArena-79K training data, reward configuration, and ColPali retriever. The Instruct variant achieves higher QA metrics on both benchmarks (EM $24.18$ vs.\ $21.85$ on MMLongBench-Doc, $47.96$ vs.\ $45.53$ on SlideVQA), while the Base variant achieves comparable retrieval recall ($60.28$ vs.\ $61.38$ on MMLongBench-Doc, $88.42$ vs.\ $86.28$ on SlideVQA). Both variants maintain low NrDup rates, indicating that the search behavior learned from DocArena transfers across base model choices.

\begin{table*}[h]
\caption{Base model comparison on MMLongBench-Doc MP and SlideVQA. Both agents are trained with the same DocArena-79K data and reward configuration.}
\vspace{-2mm}
\resizebox{\textwidth}{!}{
\begin{tabular}{c|cccc|ccc|cccc|ccc}
\hline
\multirow{3}{*}{Base Model}
& \multicolumn{7}{c|}{MMLongBench-Doc (MP)}
& \multicolumn{7}{c}{SlideVQA (MP)} \\
\cline{2-15}
& \multicolumn{4}{c|}{Retrieval}
& \multicolumn{3}{c|}{QA Metrics}
& \multicolumn{4}{c|}{Retrieval}
& \multicolumn{3}{c}{QA Metrics} \\
\cline{2-15}
& Recall $\uparrow$ & Precision $\uparrow$ & F1 $\uparrow$ & NrDup $\downarrow$
& EM $\uparrow$ & Model $\uparrow$ & PNLS $\uparrow$
& Recall $\uparrow$ & Precision $\uparrow$ & F1 $\uparrow$ & NrDup $\downarrow$
& EM $\uparrow$ & Model $\uparrow$ & PNLS $\uparrow$ \\
\hline
Qwen2.5-7B
& 60.28 & 31.36 & 37.91 & 2.10
& 21.85 & 17.34 & 29.74
& 88.42 & 51.83 & 61.22 & 0.95
& 45.53 & 40.93 & 49.11 \\
Qwen2.5-7B-Instruct
& 61.38 & 33.76 & 39.68 & 2.97
& 24.18 & 19.43 & 33.12
& 86.28 & 59.78 & 66.95 & 21.21
& 47.96 & 42.43 & 52.18 \\
\hline
\end{tabular}
}\vspace{-5mm}
\label{tab:base_model}
\end{table*}

\subsection{NrDup Metric Analysis}\label{sec:nrdup_analysis}

The Near-Duplicate Rate (NrDup) measures the percentage of multi-round queries in which at least one pair of search queries is a near-duplicate. For a given sample with search queries $\{q_1, \ldots, q_T\}$ where $T \geq 2$, we tokenize each query by lowercasing and splitting on non-alphanumeric characters, producing token sets $T_i$. A sample is flagged as containing a near-duplicate if $\exists\, (i, j),\, i \neq j$ such that $J(T_i, T_j) > 0.8$, where $J$ denotes the Jaccard similarity $|T_i \cap T_j| / |T_i \cup T_j|$. The NrDup rate is:
\begin{equation}
  \text{NrDup} = \frac{|\{s \in \mathcal{S}_{\geq 2} : \exists\, (i,j),\, J(T_i^s, T_j^s) > 0.8 \}|}{|\mathcal{S}_{\geq 2}|} \times 100\%,
\end{equation}
where $\mathcal{S}_{\geq 2}$ is the set of samples with at least two search rounds. Algorithm~\ref{alg:nrdup} provides the pseudocode.

\begin{algorithm}[h]
\caption{Near-Duplicate Rate (NrDup) Computation}\label{alg:nrdup}
\scalebox{0.75}{\begin{minipage}{1.33\columnwidth}
\KwIn{Evaluation set $\mathcal{S}$, threshold $\tau = 0.8$}
$n_{\text{dup}} \gets 0$,\; $n_{\text{multi}} \gets 0$\;
\ForEach{sample $s \in \mathcal{S}$}{
    $\{q_1, \ldots, q_T\} \gets \text{SearchQueries}(s)$\;
    \If{$T < 2$}{\textbf{continue}\;}
    $n_{\text{multi}} \gets n_{\text{multi}} + 1$\;
    $\text{flagged} \gets \textsc{False}$\;
    \ForEach{pair $(i, j)$ with $1 \leq i < j \leq T$}{
        $T_i \gets \text{Tokenize}(q_i)$,\; $T_j \gets \text{Tokenize}(q_j)$\; \Comment{lowercase, split on non-alphanumeric}
        \If{$|T_i \cap T_j| / |T_i \cup T_j| > \tau$}{
            $\text{flagged} \gets \textsc{True}$\;
            \textbf{break}\;
        }
    }
    \If{$\text{flagged}$}{$n_{\text{dup}} \gets n_{\text{dup}} + 1$\;}
}
\KwRet{$n_{\text{dup}} / n_{\text{multi}} \times 100\%$}
\end{minipage}}
\end{algorithm}

The NrDup metric captures different failure modes depending on the retrieval scenario. In single-page settings, where the target is one evidence page, a high NrDup indicates that the agent issues repetitive queries across turns without varying its search strategy. In multi-page settings, where answering requires retrieving multiple evidence pages, a high NrDup further indicates that the agent fails to formulate sufficiently distinct queries to cover different evidence pages. In both settings, a low NrDup reflects that the agent has learned adaptive, turn-aware search behavior.

%
%
\bibliographystyle{splncs04}
\bibliography{main}
\end{document}